%% file: main.tex
\documentclass{article}

\PassOptionsToPackage{numbers, compress}{natbib}

\usepackage[final]{neurips_2024}

\usepackage[utf8]{inputenc} 
\usepackage[T1]{fontenc}    
\usepackage{hyperref}       
\usepackage{url}            
\usepackage{booktabs}       
\usepackage{amsfonts}       
\usepackage{nicefrac}       
\usepackage{microtype}      
\usepackage{xcolor}         

\usepackage{amsmath}
\usepackage{amssymb}
\usepackage{mathtools}
\usepackage{amsthm}
\usepackage{mathabx}
\usepackage{multirow}  
\usepackage{bm}
\usepackage{comment}
\usepackage{algorithm}
\usepackage{algorithmic}
\usepackage{enumitem}

\usepackage{microtype}
\usepackage{graphicx}
\usepackage{subfigure}
\usepackage{booktabs} 
\usepackage{wrapfig} 

\usepackage{float}
\newfloat{algorithm}{t}{lop}

\theoremstyle{plain}
\newtheorem{theorem}{Theorem}[section]

\newtheorem{lemma}[theorem]{Lemma}

\theoremstyle{definition}
\newtheorem{definition}[theorem]{Definition}

\theoremstyle{remark}

\title{Bayesian Optimization of Functions \\ over Node Subsets in Graphs}

\author{%
  Huidong Liang \hspace{10 pt} Xingchen Wan\thanks{Now at Google.} \hspace{10 pt} Xiaowen Dong \vspace{0.05 cm} \\  
  Department of Engineering Science, University of Oxford \vspace{0.02 cm} \\ 
  \texttt{\{huidong.liang,xiaowen.dong\}@eng.ox.ac.uk, xingchenw@google.com} \\ 
}

\begin{document}

\maketitle

\input{0Abstract}
\input{1Introduction}
\input{2Preliminaries}
\input{3Method}
\input{4Experiments}
\input{5Conclusion}


\section*{Acknowledgment}
H.L. is funded by the ESRC Grand Union Doctoral Training Partnership and the Oxford-Man Institute of Quantitative Finance. X.D. acknowledges support from the EPSRC (EP/T023333/1). The authors thank members of the Machine Learning Research Group and Oxford-Man Institute of Quantitative Finance for discussing initial ideas and experiments.

\bibliography{ref}
\bibliographystyle{neurips_2024}
\clearpage 
\newpage
\appendix

\input{10RelatedWork}

\input{7ExperimentDetails}

\input{9Complexity}

\input{8KernelDetails}
\input{11Behaviour}

\input{12COMBO}
\input{13LargeGraph}
\input{14NoisySetting}
\input{15AblationStudies}



\end{document}

%% file: 0Abstract.tex
\begin{abstract}
We address the problem of optimizing over functions defined on node subsets in a graph.  The optimization of such functions is often a non-trivial task given their combinatorial, black-box and expensive-to-evaluate nature. Although various algorithms have been introduced in the literature, most are either task-specific or computationally inefficient and only utilize information about the graph structure without considering the characteristics of the function. To address these limitations, we utilize Bayesian Optimization (BO), a sample-efficient black-box solver, and propose a novel framework for combinatorial optimization on graphs. More specifically, we map each $k$-node subset in the original graph to a node in a new combinatorial graph and adopt a local modeling approach to efficiently traverse the latter graph by progressively sampling its subgraphs using a recursive algorithm. Extensive experiments under both synthetic and real-world setups demonstrate the effectiveness of the proposed BO framework on various types of graphs and optimization tasks, where its behavior is analyzed in detail with ablation studies. The experiment code can be found at \href{https://github.com/LeonResearch/GraphComBO.git}{github.com/LeonResearch/GraphComBO}.
\end{abstract}

%% file: 1Introduction.tex
\section{Introduction}
In the analysis and optimization of transportation, social, and epidemiological networks, one is often interested in finding a node subset that leads to the maximization of a utility.
For example, incentivizing an initial set of users in a social network such that it leads to the maximum adoption of certain products; protecting a set of key individuals in an epidemiological contact network such that it maximally slows down the transmission of disease; identifying the most vulnerable junctions in a power grid or a road network such that interventions can be made to improve the resilience of these infrastructure networks.

The scenarios described above can be mathematically formulated as optimizing over a utility function defined on node subsets in a graph, which is a non-trivial task for several reasons. First, most conventional optimization algorithms are designed for continuous space and are hence not directly applicable to functions defined on discrete domains such as graphs. Second, optimizing over a $k$-node subset leads to a large search space even for moderate graphs, which are not even fully observable in certain scenarios (e.g. offline social networks). 
Finally, the objective functions are usually black-box and expensive to evaluate in many applications, such as the outcome of a diffusion process on the network \cite{zhang2016dynamics} or the output of a graph neural network \cite{wan2021adversarial}, making sample-efficient queries a necessary requirement.

Assuming the graph structure is fully available, the optimization task described above shares similarities with those encountered in the literature on network-based diffusion. In that literature, greedy algorithms \cite{kempe2003maximizing,leskovec2007cost,chen2016robust} have been widely used to select a subset of nodes that maximizes a utility function, for example in the context of influence maximization \cite{sun2011survey} 
or source identification \cite{jiang2016identifying}. However, as the underlying functions often require calculating expectations over a large number of simulations (e.g. the expected number of eventual infections from an epidemic process), such algorithms often become extremely time-consuming as the evaluation time for each diffusion process increases \cite{arora2017debunking}. To relieve the inefficiency in computation, proxy-based methods, such as PageRank \cite{brin1998anatomy}, generalized random walks \cite{de2014role}, and DomiRank \cite{engsig2024domirank}, are often used in practice to rank the importance of nodes. However, such methods completely ignore the underlying function and require full knowledge of the graph structure beforehand. Finally, most methods mentioned above are task-specific, and the one designed for a specific diffusion process usually does not generalize well to another.

In this paper, we consider the challenging optimization setting for black-box functions on node subsets, where the underlying graph structure is not fully observable and can only be incrementally revealed by queries on the fly. To facilitate this setting, we propose a novel strategy to conduct the search in a combinatorial graph space (termed ``combo-graph'') in which each node corresponds to a $k$-node subset in the original (unknown) graph. The original problem is thus turned into optimization over a function on the combo-graph, where each node value is the utility of the corresponding subset. Traditional graph-traversing methods, such as breadth-first search (BFS) or depth-first search (DFS), may not work well in this case due to the exceedingly large search space and their lack of capability to exploit the behavior of the underlying function. Bayesian optimization (BO), a sample-efficient black-box solver for optimizing expensive functions via surrogate modeling of its behavior, presents an appealing alternative.

\textbf{Contributions.} 
We propose a novel Bayesian optimization framework for optimizing black-box functions defined on node subsets in a generic and potentially unknown graph. 
Our main contributions are as follows.
To the best of our knowledge, this is the first time BO has been applied to such a challenging optimization setting. Our framework consists of constructing the aforementioned combo-graph, and traversing this combo-graph by moving around a combo-subgraph sampled by a recursive algorithm. 
Notably, the proposed framework is function-agnostic and applies to any expensive combinatorial optimization problem on graphs. 
We validate the proposed framework on various graphs with different underlying functions under both synthetic and real-world settings, and demonstrate its superiority over a number of baselines. We further analyze its behavior with detailed ablation studies. Overall, this work opens new paths of research for important optimization problems in network-based settings with real-world implications.

%% file: 2Preliminaries.tex
\section{Preliminaries}\label{sec preliminaries}
BO \cite{mockus1989bayesian, garnett2023bayesian} is a gradient-free optimization algorithm that aims to find the global optimal point ${x}^*$ of a back-box function $f: \mathcal{X}\rightarrow\mathbb{R}$ over the search space $\mathcal{X}$, which, in the case of maximization, can be written as ${x}^* = \text{arg}\max_{x\in\mathcal{X}}f(x)$. To efficiently search for the optimum of expensive-to-evaluate functions, BO first builds a \textit{surrogate model} based on existing observations to predict the function values and their uncertainties over the search space $\mathcal{X}$, then utilizes an \textit{acquisition function} to decide the next location for evaluation.

\paragraph{Surrogate model.} One of the most common surrogates used in BO literature \cite{shahriari2015taking} is the \textit{Gaussian Processes} model: $f(x) \sim \mathcal{GP}(m(x), k(x, x'))$, in which $m(x)$ is the mean function (often set to a constant $\bf 0$ vector) and $k(x, x') $ is a pre-specified covariance function that measures the similarity between data point pairs. With a training set $\mathcal{D}_t = \{ {\bm x}_{1:t}, \, {\bm y}_{1:t} \}$ of $t$ observations, the posterior distribution of $f(x_{t+1})$ for a new location $x_{t+1}$ can be analytically computed from the Gaussian conditioning rule, where the mean is given by $\mathcal{\mu}(x_{t+1}|\mathcal{D}_t) =  {\bf k}(x_{t+1}, {\bf X}_{1:t}){\bf K}^{-1}_{1:t}{\bf y}_{1:t}$ with covariance $k(x_{t+1}, x'_{t+1}|\mathcal{D}_t) = k(x_{t+1},x'_{t+1}) - {\bf k}(x_{t+1}, {\bf X}_{1:t}){\bf K}^{-1}_{1:t}{\bf k}({\bf X}_{1:t}, x'_{t+1})$. Note that the computational cost for ${\bf K}^{-1}_{1:t}$ is at $\mathcal{O}(t^3)$, which largely restricts the efficiency of $\mathcal{GP}$ when training on large datasets and therefore often requires a local modeling approach.

\paragraph{Acquisition function.} Based on the predictive posterior distribution, an acquisition function will be applied to balance the exploration-exploitation trade-off via optimizing under uncertainty. For example, the \textit{Expected Improvement} \cite{mockus1998application, jones1998efficient}, defined as $\text{EI}_{1:t}(x') = \mathbb{E}\big[ [f(x') - f(x^*_{1:t})]^{+} \big]$ with $[ \alpha ]^+ = \max(\alpha, 0)$ and $x_{t}^* = \text{arg}\max_{x_i \in x_{1:t}} f(x_i)$, measures the expected improvement based on the current best query. Then, the next query location is chosen as $x_{t+1} = \text{arg}\max_{x' \in \mathcal{X}\setminus\{x_i\}_{i=1}^t} \text{EI}_{1:t}(x')$, and the result $\{x_{t+1}, y_{t+1}\}$ will be appended to the visited set $\mathcal{D}_t$. The algorithm will repeat these steps until the stopping criteria are triggered at a certain iteration $T$, and we report $x_{T}^* = \text{arg}\max_{x_i \in x_{1:T}} f(x_i)$ as the final result.

\begin{figure*}[t]
    \begin{center}
        \centerline{\includegraphics[width=\textwidth]{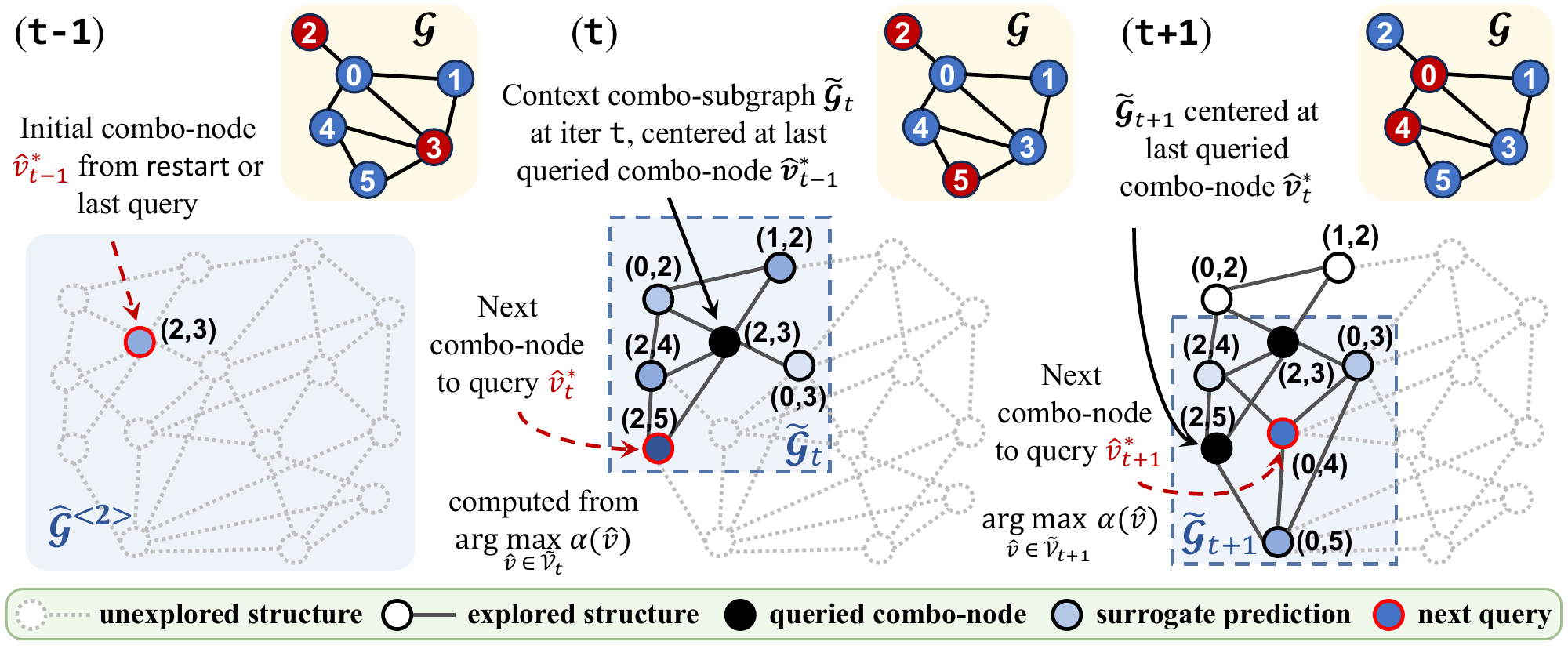}}
        \vspace{- 0.1 cm}
        \caption{Demonstration of how the proposed framework traverses the combinatorial graph $\hat{\mathcal{G}}^{<k>}$ introduced in \S\ref{sec combo-graph} with an exemplar original graph $\mathcal{G}$ of $6$ nodes and a subset size of $k=2$. At iteration \texttt{t}, we first construct a local combo-subgraph $\Tilde{\mathcal{G}}_t = \{\Tilde{\mathcal{V}}_t, \Tilde{\mathcal{E}}_t\}$ of size $Q$=6 using Algorithm~\ref{alg Combo-subgraph} (\S\ref{sec combo-subgraph}), which is centred at combo-node $\hat{v}^*_{t-1}$ from last iteration \texttt{t-1} or initialization. Next, a $\mathcal{GP}$ surrogate is fitted on $\Tilde{\mathcal{G}}_t$ with queried combo-nodes inside $\Tilde{\mathcal{G}}_t$ being the training set. The next query location is then selected as the combo-node that maximizes the acquisition function $\hat{v}^*_t = \arg\max_{\hat{v} \in \Tilde{\mathcal{V}}_t} \alpha (\hat{v})$. If queried values $f(\hat{v}^*_t) \geq f(\hat{v}^*_{t-1})$, the next combo-subgraph $\Tilde{\mathcal{G}}_{t+1}$ will be re-sampled at a new center $\hat{v}^*_t$, or otherwise remain the same. Finally, we repeat the previous process to obtain a new query location for the next iteration \texttt{t+1}, and the search continues until stopping criteria are triggered.} \label{fig structure}
    \end{center}
    \vspace{- 0.3 cm}
\end{figure*}

%% file: 3Method.tex
\section{BO of Functions over Node Subsets in Graphs}
\paragraph{Settings and challenges.}
Following the notations in \S\ref{sec preliminaries}, we formally introduce the proposed Bayesian optimization framework for black-box functions over node subsets in graphs, termed \textbf{\textit{GraphComBO}}. The goal of the problem is to find the global optimal $k$-node subset $\mathcal{S}^*$ of a black-box function $f(\mathcal{S})$ over the search space of all possible $k$-node subsets $\binom{\mathcal{V}}{k}$ on a generic graph $\mathcal{G}=\{\mathcal{V}, \mathcal{E}\}$, which, in the case of maximization, can be expressed as $\mathcal{S}^* = \arg\max_{\mathcal{S} \in \binom{\mathcal{V}}{k} }f(\mathcal{S})$. Under noisy settings, we may only observe $y = f(x) + \epsilon$, with $\epsilon \sim N(0, \sigma_\epsilon^2)$ being the noise term. For simplicity, we focus on \textit{undirected} and \textit{unweighted} graphs where the adjacency matrix $\bf A$ is symmetric and contains binary elements. As $f$ is often expensive to evaluate in practice, we wish to optimize the objective in a query-efficient manner within a limited number of evaluations $T$, and report the best configuration among them as the final solution: $\mathcal{S}^*_{T} = \arg\max_{\mathcal{S}_i \in \{\mathcal{S}_i\}_{i=1}^T }f(\mathcal{S}_i)$. 

Despite BO's appeals in optimizing such functions, we observe the following challenges when designing effective algorithms for combinatorial problems on graphs:

\begin{enumerate}[leftmargin=17pt, itemsep=0pt]
    \vspace{-3 pt}
    \item \textbf{Structural combinatorial space.} Unlike classical combinatorial optimization in the discrete space, the combination of nodes (a node subset) inherits structural information from the underlying graph, which needs to be properly encoded into the combinatorial search space. In addition, an appropriate similarity measure between node-subset pairs is also required to capture such inherent structural information when building the surrogate model.
    \item \textbf{Imperfect knowledge of graph structures.} As the complete structure of real-world graphs may be expensive or even impossible to acquire (e.g. a gradually evolving social network), any prospectus optimization algorithm needs to handle the situation where the graph structure is only revealed incrementally.
    \item \textbf{Local approach while combining distant nodes.} As the massive size $\binom{|\mathcal{V}|}{k}$ of the combinatorial space often makes global optimization unattainable, an effective local modeling approach is needed to efficiently traverse the graph. However, as the optimal subset usually consists of nodes that are far away from each other (e.g., the optimal locations of hospitals in a city network), it is critical to maintain the flexibility of selecting distant nodes when considering a local context.
\end{enumerate}

In the following sections, we will discuss how the proposed GraphComBO addresses these challenges, where an overview of the framework can be found in Figure \ref{fig structure}. 

\subsection{The Combinatorial Graph for Node Subsets} \label{sec combo-graph}
\begin{wrapfigure}{r}{0.5\textwidth}
    \vskip -0.4 cm
    \centering
    \centerline{\includegraphics[width=\linewidth]{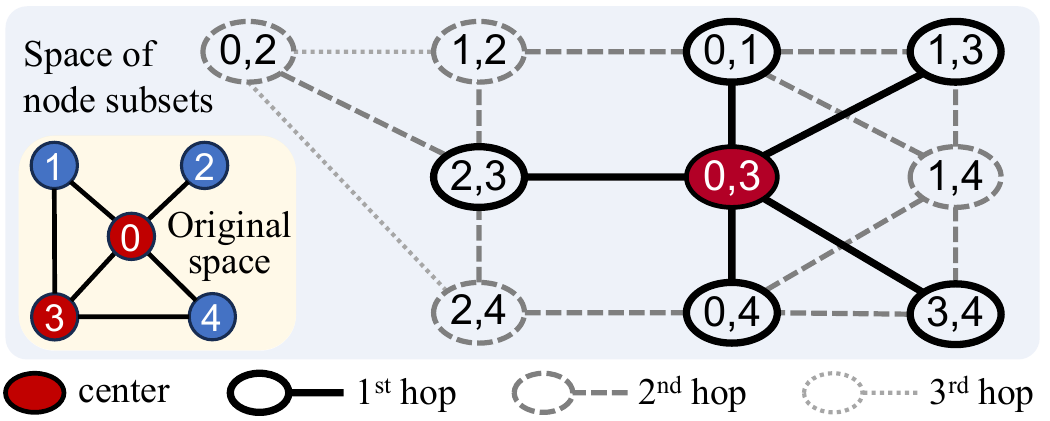}}
    \vskip -0.1 cm
    \caption{Illustration of a combinatorial graph $\hat{\mathcal{G}}^{<2>}$ constructed by the recursive combo-subgraph sampling (Algorithm \ref{alg Combo-subgraph}).} \label{fig combo-subgraph}
    \vskip -0.5 cm
\end{wrapfigure}

Inspired by the graph Cartesian product that projects multiple ``subgraphs'' into a combinatorial graph, we introduce a combinatorial graph (denoted as \textbf{\textit{combo-graph}}) tailored for node subsets on a single generic graph with an intuitive example demonstrated in Figure \ref{fig combo-subgraph}.
\begin{definition}\label{def combo-graph}
    The combinatorial operation for $k$-node subsets in an underlying graph $\mathcal{G} = (\mathcal{V}, \mathcal{E})$ is given by:
    \begin{equation}
        \hat{\mathcal{G}}^{<k>} = \Box_{i=1}^k \mathcal{G},
    \end{equation}
    which leads to a combo-graph $\hat{\mathcal{G}}^{<k>} = \{\hat{\mathcal{V}}, \hat{\mathcal{E}}\}$ of size $|\hat{\mathcal{V}}| = \binom{|\mathcal{V}|}{k}$ with each combo-node $\hat{v}_i = (v_i^{(1)}, v_i^{(2)}, ..., v_i^{(k)}) \in \hat{\mathcal{V}}$ being a $k$-node subset from the underlying graph $\mathcal{G}$ without replacement. The combo-edges $\hat{\mathcal{E}}$  in the combo-graph are defined in the following way: assume $\hat{v}_1 = (v_1^{(1)}, v_1^{(2)}, ..., v_1^{(k)})$ and $\hat{v}_2 = (v_2^{(1)}, v_2^{(2)}, ..., v_2^{(k)})$ are two arbitrary combo-nodes in the combo-graph $\mathcal{G}^{<k>}$, then $(\hat{v}_1, \hat{v}_2) \in \hat{\mathcal{E}}$ iff $\,  \exists \, j$ such that $\forall \, i \neq j$, $v_1^{(i)} = v_2^{(i)}$ and $(v_1^{(j)} , v_2^{(j)}) \in \mathcal{E}$.
\end{definition}

Intuitively, this means that in the combo-graph, two combo-nodes are adjacent if and only if they have exactly $1$ element (i.e. node from the original graph) in difference and the two different elements are neighbors in the original graph. Note that as $k$ shrinks to $1$, the combo-graph reduces to the underlying graph.

Nevertheless, as the combo-graph size $\binom{|\mathcal{V}|}{k}$ is often too large in practice, building the surrogate and making predictions at a global scale is usually unrealistic. A sensible alternative would be adopting a commonly used local modeling approach \cite{eriksson2019scalable} and then gradually moving around the ``window'' guided by the surrogate predictions. Unlike classical continuous space, constructing local regions on the combo-graph is not straightforward. Next, we will discuss two properties of the proposed combo-graph, which enable us to practically employ local modeling by sampling subgraphs for tractable optimization. Reads are also referred to \S\ref{app proof} for proofs of the following lemmas.

\begin{lemma}\label{lemma hop} \vskip 0.1 cm
    In the proposed combo-graph, at most $\ell$ elements in the subset will be changed between any two combo-nodes that are $\ell$-hop away.
\end{lemma} \vskip -0.1 cm

This implies that when considering an $\ell$-hop ego-subgraph centered at an arbitrary combo-node $\hat{v}$ on the combo-graph, we are effectively exploring the $\ell$-hop neighbors of elements in $\hat{v}$ in the original graph. Since such operation requires no prior knowledge of the other part of the original graph, we are then able to gradually reveal its structure by moving around the focal combo-node, and hence handling the situation of optimizing over node subsets on an incomplete or even unknown graph.

\begin{wrapfigure}{R}{0.5\textwidth} 
    \vspace{-0.6 cm}
    \begin{minipage}{\linewidth}
        \begin{algorithm}[H]
            \caption{Recursive combo-subgraph sampling} \label{alg Combo-subgraph}
            \textbf{Input}: Original (unknown) graph $\mathcal{G}$; The focal combo-node $\hat{v}^*$; Combo-subgraph size $Q$; Max neighbor hop $\ell_{\max}$. \\
            \textbf{Initialize}: An combo-subgraph $\Tilde{\mathcal{G}} = \{\Tilde{\mathcal{V}}, \Tilde{\mathcal{E}} \}$ with $\Tilde{\mathcal{V}} = \{\hat{v}^*\}$ $\Tilde{\mathcal{E}}=\emptyset$; Starting hop $\ell \leftarrow 1$; Set of newly found combo-nodes $\Tilde{\mathcal{V}}_{new} \leftarrow \{\hat{v}^*\}$.\\
            \textbf{Define}: \texttt{Recursive\_Sampler}($\mathcal{G}, \Tilde{\mathcal{G}}, \Tilde{\mathcal{V}}_{new}, Q, \ell $)\\
            \vspace{-1.2em}
            \begin{algorithmic} [1] 
                \FOR{$\hat{v}$ in $\Tilde{\mathcal{V}}_{new}$}
                    \FOR {$v$ in $\hat{v}$}
                        \STATE Reveal the neighbors $\mathcal{N}(v)$ of $v$ in $\mathcal{G}$.
                        \FOR{$v'$ in $\mathcal{N}(v) \cap \hat{v}$ \textbf{parallelly}}
                            \STATE Generate a new combo-node by $\texttt{CONCAT} ([v', \hat{v} \setminus v$]) and then create a combo-edge by connecting it to $\hat{v}$.
                        \ENDFOR
                        \STATE Update $\{ \Tilde{\mathcal{V}}, \Tilde{\mathcal{E}} \}$ in $\Tilde{\mathcal{G}}$.
                    \ENDFOR
                    \IF{$|\Tilde{\mathcal{V}}| > Q$ or $\ell > \ell_{\max}$}
                        \STATE Randomly drop the extra combo-nodes. 
                        \STATE \textbf{return} The combo-subgraph $\Tilde{\mathcal{G}}$ of size $Q$.
                    \ENDIF
                \ENDFOR
                \STATE Update $\Tilde{\mathcal{V}}_{new} \leftarrow  \Tilde{\mathcal{V}}\setminus\Tilde{\mathcal{V}}_{new}$; $\ell \leftarrow \ell + 1$
                \STATE \textbf{return} \texttt{Recursive\_Sampler}($\mathcal{G}, \Tilde{\mathcal{G}}, \Tilde{\mathcal{V}}_{new}, Q, \ell $)
            \end{algorithmic}
        \end{algorithm}
    \end{minipage}
    \vspace{-1.5 cm}
\end{wrapfigure}

\begin{lemma} \label{lemma linear} \vskip 0.1 cm
    The degree of combo-node $\hat{v}_i$ increases linearly with $k$ and is maximized by the subset of nodes with top $k$ degrees: $\deg(\hat{v}_i) = \sum_{j=1}^k | \mathcal{N}(v_i^{(j)}) \setminus \{v_i^{(j')}\}_{j'\neq j}^k |$.
\end{lemma} \vskip -0.1 cm

Therefore, the size of the above ego-subgraph only needs to increase linearly with $k$ to cover the first hop combo-neighbors. These two properties together make the construction of local combo-subgraphs feasible, and we introduce a sampling algorithm in the next section that recursively finds combo-nodes and combo-edges for a combo-subgraph (denoted as $\Tilde{\mathcal{G}}$) given a focal combo-node.

\paragraph{Recursive combo-subgraph sampling.} \label{sec combo-subgraph}
As illustrated in Algorithm \ref{alg Combo-subgraph} and Figure \ref{fig combo-subgraph}, our goal is to construct an ego-subgraph $\Tilde{\mathcal{G}}$ of size $Q$ from the underlying graph $\mathcal{G}$, centered at a given combo-node $\hat{v}^*$ with maximum hop $\ell_{\max}$. The algorithm initializes $\Tilde{\mathcal{G}} = \{\Tilde{\mathcal{V}}, \Tilde{\mathcal{E}} \}$ with only $\hat{v}^*$ and then loop through neighbors of each element node $v \in \hat{v}^*$. If a neighbor $\mathcal{N}_i(v)$ of $v$ is not in $\hat{v}^*$ (i.e. ensuring no repetition in the subset), a new combo-node $\hat{v}'$ will be created by substituting $\mathcal{N}_i(v)$ with $v$ in $\hat{v}^*$, and a combo-edge will be accordingly created by connecting $\hat{v}'$ to $\hat{v}^*$. As a result, after finding the combo-neighbors of $\hat{v}^*$ at hop $\ell=1$, $\Tilde{\mathcal{G}}$ becomes a star-network at center $\hat{v}^*$. We will then repeat the above procedures (i.e. star-sampling) for every newly found combo-node to find their combo-neighbors at hop $\ell+1$ (which meanwhile also finds the edges among combo-nodes within the previous hop $\ell$), until the subgraph size limit $Q$ or the maximum hop $\ell_{\max}$ is reached. 

By constructing the combo-graph and sampling subgraphs from it, we can efficiently traverse the combinatorial space by progressively moving around the combo-subgraph center while preserving diversified combinations of distant nodes under a local modeling approach, which will be discussed in the following section.

\subsection{Graph Gaussian Processes Surrogate}\label{sec GP}
After constructing the combo-subgraph, we can build a surrogate model for the expensive underlying function on this local region with graph Gaussian Processes ($\mathcal{GP}$). Specifically, we consider the normalized graph Laplacian $\Tilde{\bf L}$: $\Tilde{\bf L} = {\bf I} - \Tilde{\bf D}^{-1/2}\Tilde{\bf A}\Tilde{\bf D}^{-1/2}$ for a combo-subgraph $\Tilde{\mathcal{G}}$, where $\Tilde{\bf A}$ is the adjacency matrix and $\Tilde{\bf D}$ is the degree matrix. Then, the eigendecomposition of the graph Laplacian matrix is given by $\Tilde{\bf L} = {\bf U} {\bm \Lambda} {\bf U}^\top$, in which ${\bm \Lambda}=\text{diag}(\lambda_1, \cdots, \lambda_n)$ are the eigenvalues sorted in ascending order and ${\bf U}=[{\bm u}_1, \cdots, {\bm u}_n]$ are their corresponding eigenvectors. Now let $i, j \in \{1, \cdots, n\}$ be two indices of combo-nodes on $\Tilde{\mathcal{G}}$, the covariance function (or \textit{kernel}) $k(\hat{v}_i, \hat{v}_j)$ between an arbitrary combo-node pair $\hat{v}_i$ and $\hat{v}_j$ can be formulated in the form of a regularization function $r(\lambda_p)$ \cite{smola2003kernels} defined on the eigenvalues $\{\lambda_p\}_{p=1}^n$:
\begin{equation}
    k(\hat{v}_i, \hat{v}_j) = \sum_{p=1}^n r^{-1}(\lambda_p){\bm u}_p[i]{\bm u}_p[j], \label{eq kernel}
\end{equation}
where ${\bm u}_p[i]$ and ${\bm u}_p[j]$ are the $i$-th and $j$-th elements in the $p$-th eigenvector ${\bm u}_p$, and $r(\lambda_p)$ is some scalar-valued function for regularization. We refer readers to Appendix~\S\ref{app kernel} for discussion on a collection of commonly used kernels on graphs under the form of Equation~\eqref{eq kernel}. 
\begin{wrapfigure}{R}{0.5\textwidth}   
    \vspace{-0.6 cm}
    \begin{minipage}{\linewidth}
        \begin{algorithm}[H]
            \caption{BO for node-subsets on graphs} \label{alg BO}
            \textbf{Input}: Original (unknown) graph $\mathcal{G}=\{\mathcal{V}, \mathcal{E}\}$; underlying function $f$ defined on $k$- node subsets $\mathcal{S} \in \binom{|\mathcal{V}|}{k}$; combo-subgraph size $Q$; \# failures threshold $\texttt{failtol}$; \# queries $T$. \vspace{0.02 cm} \\
            \textbf{Objective}: Find $\mathcal{S}^*_{T} = \arg\max_{\mathcal{S}_i \in \{\mathcal{S}_i\}_{i=1}^T }f(\mathcal{S}_i)$. \\
            \textbf{Initialize}: Set initial training set $\mathcal{D}_0 \leftarrow \emptyset$ and queried set $\mathcal{O}_0 \leftarrow \emptyset$; Set restart status $\texttt{restart} \leftarrow \texttt{True}$; Set counter of non-improvement tolerance $F \leftarrow 0$. Use initial \texttt{start\_location} $\hat{v}_0$ if applicable, and specify the \texttt{restart\_method} for \texttt{restart}. \\
            \vspace{-1.2em}
            \begin{algorithmic}[1] 
                \FOR{$t=1,\cdots,T$}
                    \IF{\texttt{restart}}
                        \STATE Re-initialize the starting location $\hat{v}_t$ using \texttt{restart\_method} and \texttt{start\_location}; Query $\hat{v}_t$ and reset the training set $\mathcal{D}_t \leftarrow (\hat{v}_t, y_t)$; Set $\texttt{restart} \leftarrow \texttt{False}$.
                    \ENDIF
                    \STATE Sample a combo-subgraph $\Tilde{\mathcal{G}}_t=\{\Tilde{\mathcal{V}}_t, \Tilde{\mathcal{E}}_t\}$  with $|\Tilde{\mathcal{V}}_t| = Q$ centered at the \textbf{best} training combo-node $\hat{v}^*_{t-1}$ from $\mathcal{D}_{t-1}[\hat{v}]$ using Algorithm~\ref{alg Combo-subgraph} in \S\ref{sec combo-subgraph}.
                    \STATE Fit the $\mathcal{GP}$ surrogate defined in \S\ref{sec GP} on $\Tilde{\mathcal{G}}_t$ by maximum likelihood, with the queried combo-nodes inside $\Tilde{\mathcal{G}}_t$ being the training set (i.e. $\mathcal{D}_{t}[\hat{v}] = \Tilde{\mathcal{V}}_t \cap \mathcal{O}_{t-1}[\hat{v}]$).
                    \STATE Optimize the acquisition function $\alpha$; Select $\hat{v}_t = \arg \max_{\hat{v} \in \Tilde{\mathcal{V}}_t} \alpha (\hat{v})$ from $\Tilde{\mathcal{G}}_t$ as the next query; Obtain the function value $y_t$ at $\hat{v}_t$.         
                    \STATE Update query set $\mathcal{O}_t \leftarrow \mathcal{O}_{t-1} \cup (\hat{v}_t, y_t)$ and training set $\mathcal{D}_t \leftarrow \mathcal{D}_{t-1} \cup (\hat{v}_t, y_t)$; Select the best training combo-node $\hat{v}^*_t = \arg \max_{\hat{v} \in \mathcal{D}_t[\hat{v}]} f(\hat{v})$.
                    \STATE Update $F \leftarrow F + 1$ \textbf{if} $\hat{v}^*_t = \hat{v}^*_{t-1}$; Set $\texttt{restart} \leftarrow \texttt{True}$ \textbf{if} $\texttt{failtol} = F$ \textbf{or} all combo-nodes in $\Tilde{\mathcal{G}}_t$ are queried.
                \ENDFOR
                \STATE \textbf{return} $\hat{v}_T^* = \arg\max_{\hat{v} \in \mathcal{O}_T[\hat{v}]} f(\hat{v})$. \vspace{-0.0 cm}
            \end{algorithmic}
        \end{algorithm}
    \end{minipage}
    \vspace{-1.5 cm}
\end{wrapfigure}

\subsection{Bayesian Optimization on the Combo-graph} \label{sec GraphComBO}
With the structural combinatorial space and techniques to sample and build surrogate models on the combo-subgraphs, we now introduce the proposed GraphComBO framework in detail. For simplicity, we consider maximization in the following paragraphs, where the overall structure can be found in Figure \ref{fig structure} with key procedures summarized in Algorithm~\ref{alg BO} and complexity discussed in Appendix \S\ref{app complexity}.

\paragraph{Combo-subgraphs as trust regions.}
As discussed earlier, performing global modeling directly for combinatorial problems is usually impractical. Thus, inspired by the \textit{trust region} method popularly used in continuous numerical optimization \cite{conn2000trust}, reinforcement learning \cite{schulman2015trust} and BO under other settings \cite{eriksson2019scalable,wan2021think}, we take a local modeling approach on the combo-graph during the BO search. Starting with a random location (i.e. a combo-node $\hat{v}_0$) or a reasonable guess from domain knowledge, a combo-subgraph $\Tilde{\mathcal{G}}_0$ will be constructed at center $\hat{v}_0$ by Algorithm \ref{alg Combo-subgraph}. We will then move around this combo-subgraph $\Tilde{\mathcal{G}}_t$ at each iteration $t$ on the combo-graph by changing its focal combo-node guided by the surrogate model and acquisition function, which will be explained shortly.

In particular, we introduce a hyperparameter $Q$ that caps the combo-subgraph size to control the computational cost for the surrogate $\mathcal{GP}$, and then use the queried combo-node inside $\Tilde{\mathcal{G}}_t$ as the training set $\mathcal{D}_t$ to fit the model (i.e. update the hyperparameters in its kernel). The acquisition function $\alpha(\hat{v})$ is then applied on the rest of unvisited combo-nodes in $\Tilde{\mathcal{G}}_t$, and we select the combo-node $\hat{v}_t = \arg\max_{\hat{v} \in \Tilde{\mathcal{V}}_t} \alpha(\hat{v})$ as the next location to query the underlying function. Here, any commonly used kernel and acquisition function are compatible with our setting, and we adopt the popular \textit{diffusion} kernel \cite{oh2019combinatorial} with \textit{Expected Improvement} acquisition \cite{jones1998efficient} in our experiments.

After querying the next location, we re-select the best-queried combo-node $\hat{v}^*_t$ in our training set $\mathcal{D}_t[\hat{v}]$ by choosing $\hat{v}^*_t = \arg \max_{\hat{v} \in \mathcal{D}_t[\hat{v}]} f(\hat{v})$,  and compare it to the previous best location $\hat{v}^*_{t-1}$. If the best-queried value improves (i.e. $f(\hat{v}^*_t) > f(\hat{v}^*_{t-1})$), the combo-subgraph in the next iteration $\Tilde{\mathcal{G}}_{t+1}$ will be resampled at this new location $\hat{v}^*_t$ with Algorithm~\ref{alg Combo-subgraph}, or otherwise remains the same as $\Tilde{\mathcal{G}}_{t}$. The search algorithm then continues until a querying budget $T$ is reached, and we report the best-queried combo-node as the final result $\hat{v}_T^* = \arg\max_{\hat{v} \in \hat{v}_{1:T}} f(\hat{v})$.

\vspace{-0.1 cm}

\paragraph{Balancing exploration and exploitation.}
Similar to the continuous domain, the exploration-exploitation trade-off is also a fundamental concern when using BO on the proposed combo-graph, and we introduce two additional techniques to strike a balance between these two matters.

1. \texttt{failtol} that controls the tolerance of ``failures'' by counting continuous non-improvement steps. Once reached, the algorithm will restart at a new location using \texttt{restart\_method}.

2. \texttt{restart\_method} that either restarts at a random combo-node, the best-visited combo-node, or the initial starting location if specified.

In addition, the combo-subgraph size $Q$, which can be viewed as the ``volume'' of the trust region under graph setting, also controls the step size of exploration. These strategies together can cohesively assist GraphComBO in adapting to various tasks. For example, a small \texttt{failtol} will encourage exploration in the combinatorial space when \texttt{restart\_method} is set to a random combo-node, which is useful when optimizing an underlying function with low graph signal smoothness \cite{dong2016learning}. By contrast, when increasing \texttt{failtol} and setting \texttt{restart\_method} to the best-queried combo-node, the algorithm will exploit more around the local optimal and is hence more suitable for smoother functions. \S\ref{app ablation} further provides an ablation study on these hyperparameters.

\paragraph{Impact of the underlying function $f$ and the subset size $k$.} \label{sec subset size} 
It is natural to expect that the interaction between the underlying function $f$ and graph structure, which relates to signal smoothness over the combo-graph, will exert a significant influence on the search performance. Specifically, the optimization is expected to be challenging either when $f$ is less correlated to the graph structure even if the latter is informative (e.g. random noise on a BA network~\cite{barabasi1999emergence} as an extreme case), or when $f$ is correlated to the graph structure but the latter is non-informative (e.g. eigenvector centrality on a Bernoulli random graph~\cite{erdos1959evolution}). In the meantime, as the subset size $k$ increases, exploration will become more expensive when using a combo-subgraph of fixed size $Q$ or fixed number of hops~$\ell_\text{max}$. Recall that in \textbf{Lemma~\ref{lemma hop}} where we state that at most $\ell$ elements will be changed between two combo-nodes that are $\ell$-hops away, it implies that more queries are required to exhaust all possible modifications of the elements in the subset when its size $k$ increases. 
Empirical findings from our experiments in \S\ref{sec exp} further corroborate these hypotheses, where we also provide detailed discussions on model behavior in \S\ref{app behaviour} and kernel performance under different levels of signal smoothness in \S\ref{app smoothness}. 

\paragraph{Relation to previous BO methods with graph settings.} While BO has been combined with graph-related settings to find the optimal {\em graph structures} such as in the literature of NAS \cite{kandasamy2018neural, oh2019combinatorial,ru2020interpretable} and graph adversarial attacks \cite{wan2021adversarial}, it remains largely under-explored for optimizing functions defined on the {\em nodes} or {\em node subsets} in the graph. Although one recent work \texttt{BayesOptG}~\cite{wan2023bayesian} considered such novel setups, it only considered functions defined on a single node, which can be viewed as a special case in our setting when $k=1$. The construction of the ``combo-graph'' in our approach shares similarity with the construction of the combinatorial graph in \texttt{COMBO}~\cite{oh2019combinatorial}; however, the problems being addressed there do not arise in a natural graph setting, and we present a more detailed discussion of the related work in Appendix~\S\ref{app related work}, together with an additional experiment for comparison in \S\ref{app COMBO}.

\begin{figure*}[t]
    \begin{center}
        \centerline{\includegraphics[width=\textwidth]{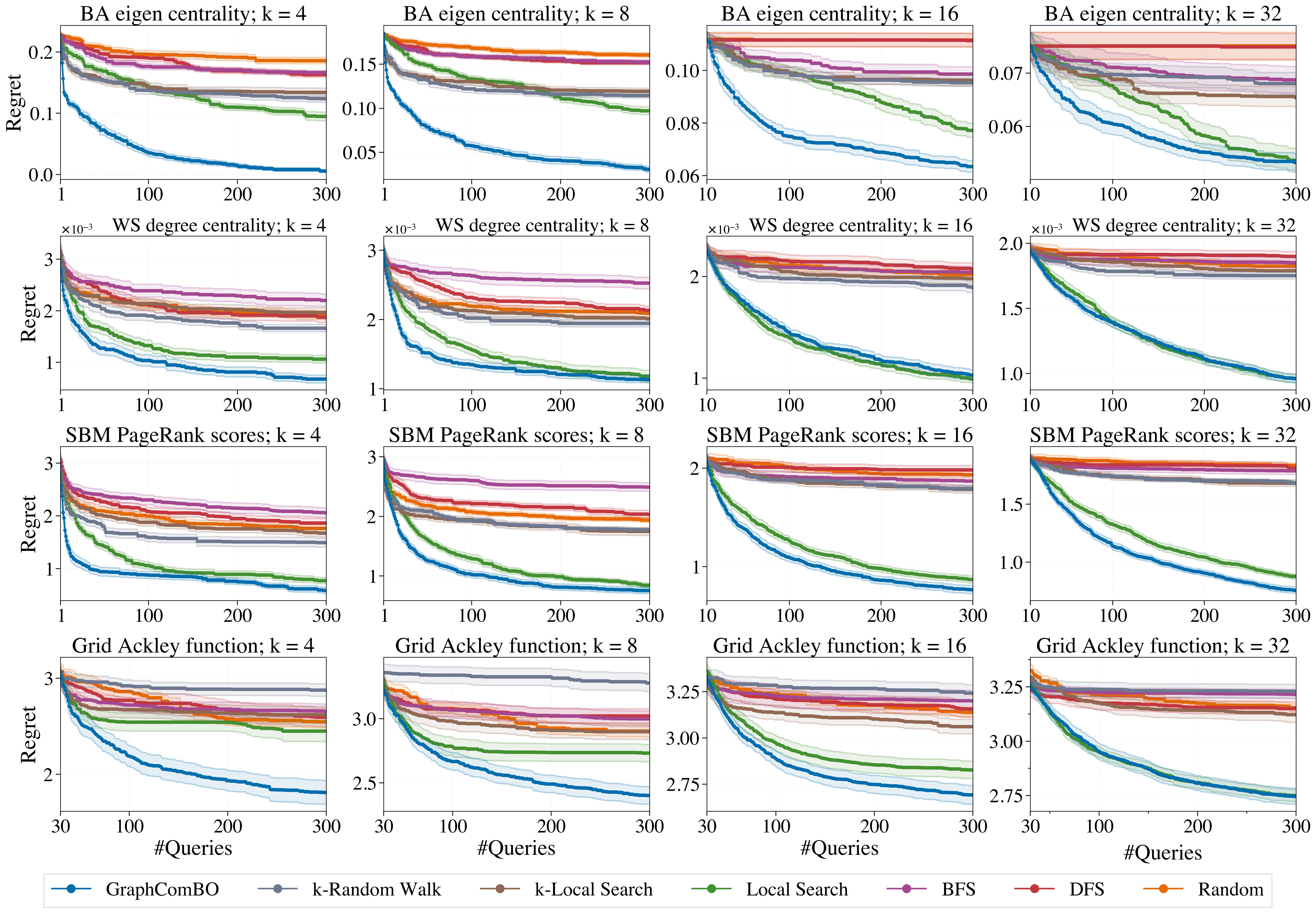}}
        \vspace{-0.2 cm}
        \caption{Results for synthetic problems on BA, WS, SBM and 2D-Grid networks with $k=[4,8,16,32]$, where \texttt{Regret} indicates the difference between ground truth and the best query so far.} \label{fig synthetic results}
    \end{center}
    \vskip -0.6 cm
\end{figure*}

%% file: 4Experiments.tex
\section{Experiments} \label{sec exp}
\paragraph{Setups.} 
We conduct comprehensive experiments on four synthetic problems and five real-world tasks to validate our proposed framework, where readers are also referred to the appendix for discussions on: \S\ref{app experiment} detailed experimental settings with task descriptions and visualizations; \S\ref{app kernel} validation of common kernels on graphs under our settings; \S\ref{app behaviour} a thorough analysis of GraphComBO's underlying behavior; and \S\ref{app ablation} ablation studies on the hyper-parameters. We closely follow the standard setups in BO literature \cite{baptista2018bayesian,eriksson2019scalable,hvarfner2022pibo}. Specifically, we query 300 times and repeat 20 times with different random seeds for each task, in which the mean and standard error of the cumulative optima are reported for all methods. For simplicity, we use a diffusion kernel~\cite{oh2019combinatorial} with automatic relevance determination and adopt \textit{Expected Improvement}~\cite{jones1998efficient} as the acquisition function to investigate subset sizes of $k=[4,8,16,32]$, where we also fix $Q=4,000$ and $\texttt{failtol}=30$ across all experiments. In addition, we also initialize the algorithm with $10$ queries using simple random walks when $k \geq 16$. 

\paragraph{Baselines.} 
As the proposed framework is a first-of-its-kind BO method for optimizing expensive and black-box functions of node subsets on generic graphs, we consider three graph-traversing algorithms that operate on the original graph: \textit{Random}, \textit{k-Random Walk} and \textit{k-Local Search}; and three algorithms on the proposed combo-graph: \textit{BFS}, \textit{DFS} and \textit{Local Search} as the baseline methods, with their details described in Appendix \S\ref{app experiment}. Notably, the local search method, which randomly queries a neighbor of the best-queried combo-node at each iteration, can be viewed as a BO method that uses a ``random'' surrogate model, and hence serves as a good indicator for GraphComBO's behavior.

\begin{figure*}[t]
    \begin{center}
        \centerline{\includegraphics[width=\textwidth]{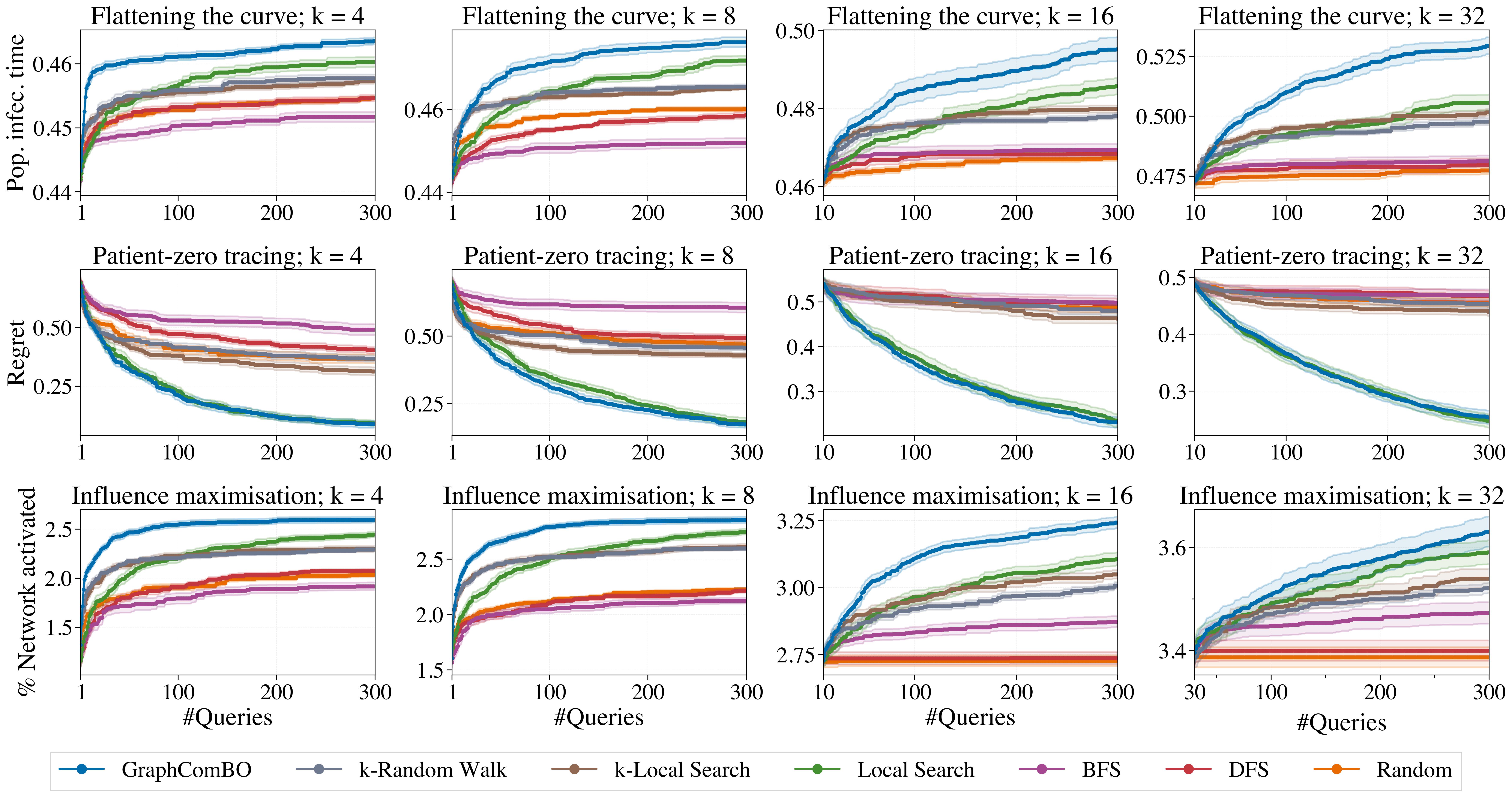}}
        \vskip -0.2 cm
        \caption{Results for flattening the curve, patient-zero tracing and influence maximization.} \label{fig SIR IC results}
    \end{center}
    \vskip -0.5 cm
\end{figure*}

\paragraph{Synthetic problems on random graphs.}\label{sec experiment synthetic}
We first validate the proposed framework on four ubiquitous random graph types with commonly used analytical underlying functions. Concretely, we consider Barab\'{a}si-Albert (BA)~\cite{barabasi1999emergence}, Watts-Strogatz (WS)~\cite{watts1998collective}, stochastic block model (SBM)~\cite{holland1983stochastic} and 2D-grid networks, where their corresponding ``base'' underlying functions are eigenvector centrality, degree centrality, PageRank~\cite{brin1998anatomy}, and Ackley function~\cite{ackley2012connectionist}, respectively. We then take the average over node values inside a subset to obtain the final underlying function, which, given the analytical setting, enables us to compute the difference (\texttt{Regret}) between the queried-best value and ground truth. The search results are presented in Figure~\ref{fig synthetic results}, where we also explained the problem settings in detail in Appendix \S\ref{app sec synthetic} and summarized the graph statistics in Table~\ref{tab statistics}.

\begin{figure*}[t]
    \begin{center}
        \centerline{\includegraphics[width=\textwidth]{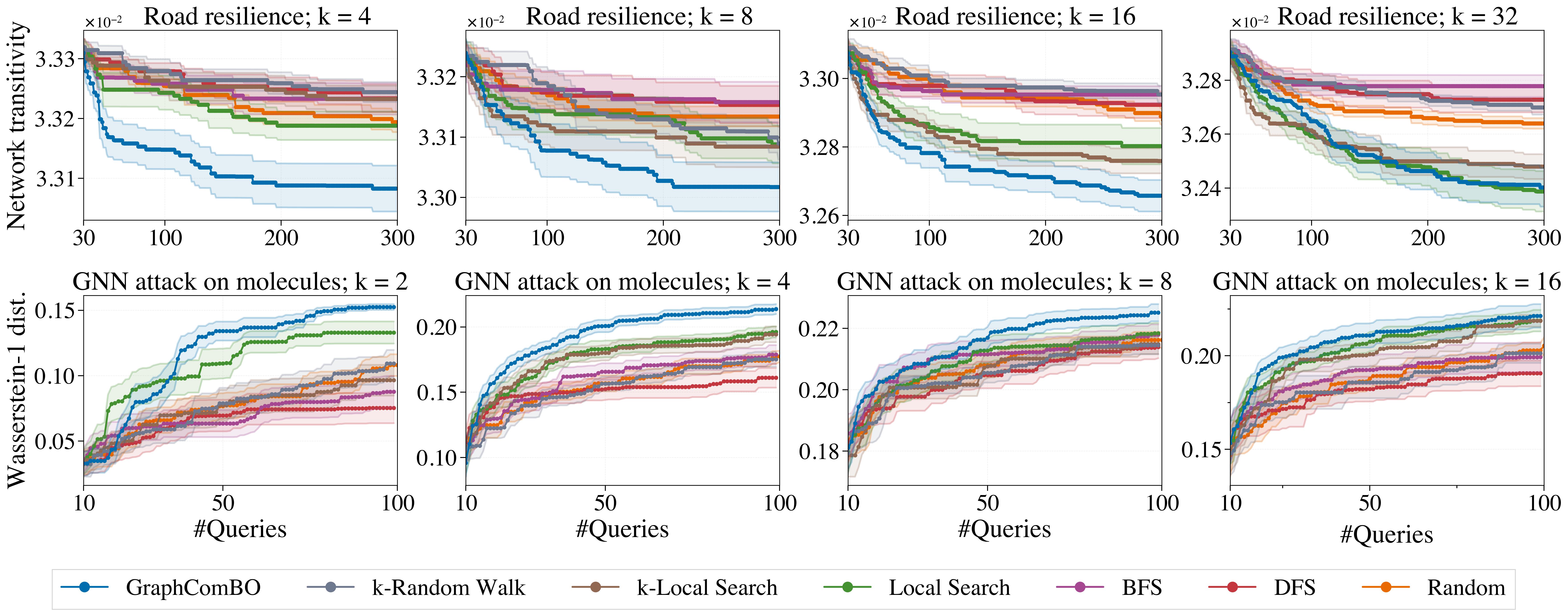}}
        \vspace{-0.1 cm}
        \caption{Results for road resilience testing and GNN attacks on molecules with edge-masking.} \label{fig road gnn results}
    \end{center}
    \vskip -0.5 cm
\end{figure*}

\paragraph{Real-world optimization tasks.} \label{sec experiment real-world}
After validation under synthetic settings, we carry out five real-world experiments on epidemic contact networks, social networks, transportation networks, and molecule graphs, where their results are presented in Figure~\ref{fig SIR IC results} and Figure~\ref{fig road gnn results}. The statistics of the underlying functions and graphs are summarized in Table~\ref{tab statistics}, where the detailed setting for each scenario is explained and visualized in \S\ref{app SIR}-\ref{app gnn}. Specifically, we consider the following tasks:
\begin{itemize}[leftmargin=17pt] 
    \item \textbf{\textit{Flattening the curve in epidemics (\S\ref{app SIR})}.}
     We adopt the widely-used SIR simulations~\cite{kermack1927contribution} on a real-world contact network with a goal of protecting $k$ nodes in the network, such that the expected time of reaching $50\%$ population infection will be maximally delayed. 
    
    \item \textbf{\textit{Identifying patient-zero in communities (\S\ref{app patient zero})}.} 
    We apply SIR on an SBM network to simulate a disease contagion across multiple communities, where the goal is to identify $k$ individuals with the earliest infection time, given the complete transmission network not known a priori. 
    
    \item \textbf{\textit{Maximizing influence on social networks (\S\ref{app IC})}.} 
    We consider the influence maximization problem on a social network~\cite{shchur2018pitfalls} with independent cascading simulations~\cite{kempe2003maximizing}, in which we aim to select the optimal $k$ nodes as the seeds (i.e. source of influence) that maximize the expected number of final influenced individuals (represented as a fraction of the network size).
    
    \item \textbf{\textit{Resilience testing on transportation networks (\S\ref{app road})}.} The objective of this task is to identify the $k$ most vulnerable roads (edges), such that their removal will lead to the maximal drop in a certain utility function measuring the operation status (estimated by network transitivity).
    
    \item \textbf{\textit{Black-box attacks on graph neural networks (\S\ref{app gnn})}.} 
    Considering a graph-level GNN pre-trained for molecule classification \cite{morris2020tudataset} with a particular input graph, we conduct a challenging black-box attack with no access to the model parameter but only a limited number of queries for its output. Our goal here is to mask $k$ edges such that the output from the victim GNN (at softmax) will be maximally perturbed from the original output, as measured by the Wasserstein distance. 
\end{itemize}

\paragraph{Discussion on results.} 
We can observe that the proposed GraphComBO framework generally outperforms all the other baselines with a clear advantage on both synthetic and real-world tasks. It is worth noting that such gain in performance seems to be diminishing as $k$ increases and, in certain scenarios, BO also tends to perform similarly to local search. While these phenomena are generally consistent with our previous hypothesis in \S\ref{sec subset size}, we further provide the following explanations to attain a better understanding of the model's underlying behavior. 
\begin{enumerate}[leftmargin=17pt, itemsep=0pt] \vskip -0.5 cm
    \item  Given a combo-subgraph with a fixed size $Q$, we tend to capture structural information in a smaller neighborhood due to the increase of combo-node degrees. First, when $k$ increases, the combo-node degree will increase linearly (as discussed in \textbf{Lemma~\ref{lemma linear}}). Second, if the synthetic underlying function has a strong positive correlation with node degree, such as eigenvector centrality, the degree of the center combo-node will increase as the search progresses. Both factors will lead to a smaller neighborhood around the focal node (in terms of shortest path distance) covered by the combo-subgraph, which in turn means more steps are required to explore beyond the current region, especially when the algorithm reaches a local optimum.

    \item As discussed in \S\ref{sec subset size}, underlying functions with low signal smoothness will negatively affect BO's performance, which partially explains the comparable results between BO and local search on WS and SBM where the graph structures are less informative, as well as on the molecule network where the underlying function involves a graph neural network and is relatively non-smooth compared to other tasks. In addition, as the kernels used in $\mathcal{GP}$ (\S\ref{sec GP}) come with an underlying assumption on function smoothness, the surrogate model will capture less signal information when fitting a less-smooth function, thus making BO behave similarly to a random model (i.e. the local search). To better support this claim, we further conduct a sensitivity analysis of the kernels to signal smoothness at different levels in Appendix \S\ref{app smoothness}.
\end{enumerate}

\paragraph{Further analysis on model behaviors.} Readers are also referred to Appendix \S\ref{app behaviour} for a more detailed behavior analysis that elaborates on the above explanations and Appendix \S\ref{app ablation} for a thorough ablation study on $Q$ and $\texttt{failtol}$. In addition, Appendix \S\ref{app COMBO} provides a comparison with COMBO~\cite{oh2019combinatorial} on small-scale networks, Appendix \S\ref{app large graph} tests our framework on a large social network OGB-arXiv with $ |\mathcal{V}| = 1.7\times10^5$, and finally Appendix \S\ref{app nosiy} discusses the framework's performance under a noisy setting where observations are corrupted at different noise levels.


%% file: 5Conclusion.tex
\section{Conclusion and Future Works} \label{app limitation}
In this work, we introduce a novel Bayesian optimization framework to optimize black-boxed functions defined on node subsets in a generic and potentially unknown graph. By constructing a tailored combinatorial graph and sampling subgraphs progressively with a recursive algorithm, we are able to traverse the combinatorial space and optimize the objective function using BO in a sample-efficient manner. Results on both synthetic and real-world experiments validate the effectiveness of the proposed framework, and we use detailed analysis to study its underlying behavior.

On the other hand, we have also identified the following limitations during our experiments, which can be explored as future directions for this line of work.

\begin{itemize}[leftmargin=17pt]
    \item As discussed in the paper, the performance of BO gradually deteriorates when the subset size $k$ increases. In this sense, some modifications are expected to better control the combinatorial explosion while preserving useful information from the underlying graph structure. 
    \item The proposed framework adopts a local modeling approach inspired by the trust region method to control the computational cost. However, we expect some improvement in BO's performance if we inject some global information (if available) into surrogate modeling, such as using some self-supervised method with a graph neural network to replace the Laplacian embedding.
    \item The current algorithm adopts a fixed strategy for hyperparameters like subgraph size $Q$ and maximum hop $\ell_{\max}$, where we believe the optimization would benefit from a more flexible design such as a self-adaptive $Q$ and $\ell_{\max}$ as the search continues.
    \item In all experiments, we assume no prior knowledge of the problem and adopt a random initialization method before the search. However, it is also an important direction to explore when a good starting location or certain characteristics of the function are available from domain knowledge.
\end{itemize}

We believe the proposed combo-graph would bring new insights to a broader community of machine learning research on graphs. While there are many potential societal consequences of our work, none of which we feel must be specifically highlighted here.

%% file: 10RelatedWork.tex
\section{Related Work}\label{app related work}
\paragraph{BO for combinatorial optimization.} Bayesian optimization has been widely applied to solving combinatorial problems with black-box and expensive-to-evaluate underlying functions~\cite{shahriari2015taking, wang2023recent}. Notably, \texttt{BOCS} \cite{baptista2018bayesian} handles the combinatorial explosion of the discrete search space by utilizing an approximate optimizer for the acquisition function, which addresses the limited scalability of common acquisition functions to
large combinatorial domains; \texttt{CoCaBo}~\cite{ru2020bayesian} further tackles the setting of mixed search space with multiple categorical variables by introducing a $\mathcal{GP}$ kernel to capture the interaction between continuous and categorical inputs; and \texttt{LADDER}~\cite{deshwal2021combining} takes a latent variable approach that first encodes the problem into a latent space via unsupervised learning and then adopts a structure-coupled kernel, which integrates both decoded structures and the latent representation for better surrogate modeling. While BO has provided a sample-efficient way for the above combinatorial optimization problems, its extension to settings with graph structures, especially when the search space itself is a generic graph, still remains largely under-explored and will be discussed in the following section.

\paragraph{BO with graphs.} Despite several works in the literature combining BO with graph-related settings, the majority of them focus on optimization over graph inputs (i.e. each configuration itself is a graph, and the goal is to optimize for graph structures). In particular, \texttt{NASBOT}~\cite{kandasamy2018neural} treats the neural network architecture as a graph structure and uses BO to perform neural architecture search, while \texttt{NAS-BOWL}~\cite{ru2020interpretable} approaches the same problem from a different perspective by using Weisfeiler-Lehman kernels with BO. Other examples include using BO for molecular graph designs \citep{korovina2020chembo}, graph adversarial attack \citep{wan2021adversarial}, and a general framework for optimizing functions on graph structures \citep{cui2021novel}. These works, however, are under a different setup compared to our current work, which aims to optimize functions defined on node subsets in a single generic and potentially unknown graph. Moreover, the above works typically seek for a global vector-embedding of the graph configuration, after which standard kernels will be applied to measure their similarity in the Euclidean space.

On the other hand, \texttt{BayesOptG}~\cite{wan2023bayesian} proposes a framework that employs BO to optimize functions defined on a single node on graphs, where the search space is a generic graph and the configurations are nodes on the graph. The similarity between two configurations is then measured between node pairs with kernels on graphs capturing structural information. As mentioned earlier, our paper is a generalization of this previous work by considering a combinatorial setting, in which \texttt{BayesOptG} can be viewed as a special case under our framework when the number of nodes in the subset is $k=1$. 

Lastly, another relevant line of works from the literature is \texttt{COMBO}~\cite{oh2019combinatorial} and its variants \cite{imani2021graph,deshwal2021mercer,ocariz2023neural}, where the underlying function is defined on a Cartesian product graph computed from $k$ small graphs. In particular, each node on this combinatorial graph represents a combination of $k$ elements from the $k$ graphs, after which kernels on graphs are leveraged to measure the similarity between nodes in the combinatorial space. Nevertheless, the combinatorial graph introduced in our work differentiates substantially from that in \texttt{COMBO}, and we emphasize the differences in the following.
\begin{enumerate}[leftmargin=20pt]
    \item From the problem setting, \texttt{COMBO} is designed for $k$-node combinations from $k$ distinct graphs $\{G_i\}_{i=1}^{k}$ (i.e. one node from each graph, which corresponds to one variable in the combinatorial optimization). The structure of these graphs and the resulting combinatorial graph is therefore pre-defined and fully available. In contrast, our work is concerned with the combination of $k$ nodes (a $k$-node subset) from a single and generic graph whose structure (and hence the structure of our combo-graph) is potentially unknown a priori.
    \item The resulting search space for \texttt{COMBO} is of dimension $\prod_{i=1}^{k}N_i$ with $N_i$ corresponding to the size of $i$th graph $G_i$, whereas in our work, the combinatorial space has a dimension $\binom{N}{k}$ with $N$ being the size of $G$. Note that in \texttt{COMBO}, even if we have $k$ identical graphs and naively calculate their Cartesian product, the resulting space $N^k$ is still not the same as our case. For instance, $(a,b)$ and $(b,a)$ are two distinct sets in \texttt{COMBO}, but they should be considered as one single set in our scenario. Besides, sets with duplicated elements (e.g., $(a,a)$ or $(a,a,b)$) are valid in \texttt{COMBO} but meaningless in our setting, which will introduce redundant computational costs.
    \item One of the main contributions of \texttt{COMBO} is that the eigendecomposition of the large combinatorial graph can be performed on the $k$ smaller graphs with Kronecker product operation. Nevertheless, it is limited to small $k$ and $N_i$ as the memory is simply not large enough to fit in an eigenbasis of size $(\prod_{i=1}^{k}N_i)^2$ even with moderate choices of $k$ and $N_i$. 
\end{enumerate}

%% file: 7ExperimentDetails.tex
\section{Experimental Details}\label{app experiment}

\begin{figure}[t]
    \begin{center}
        \centerline{\includegraphics[width=\textwidth]{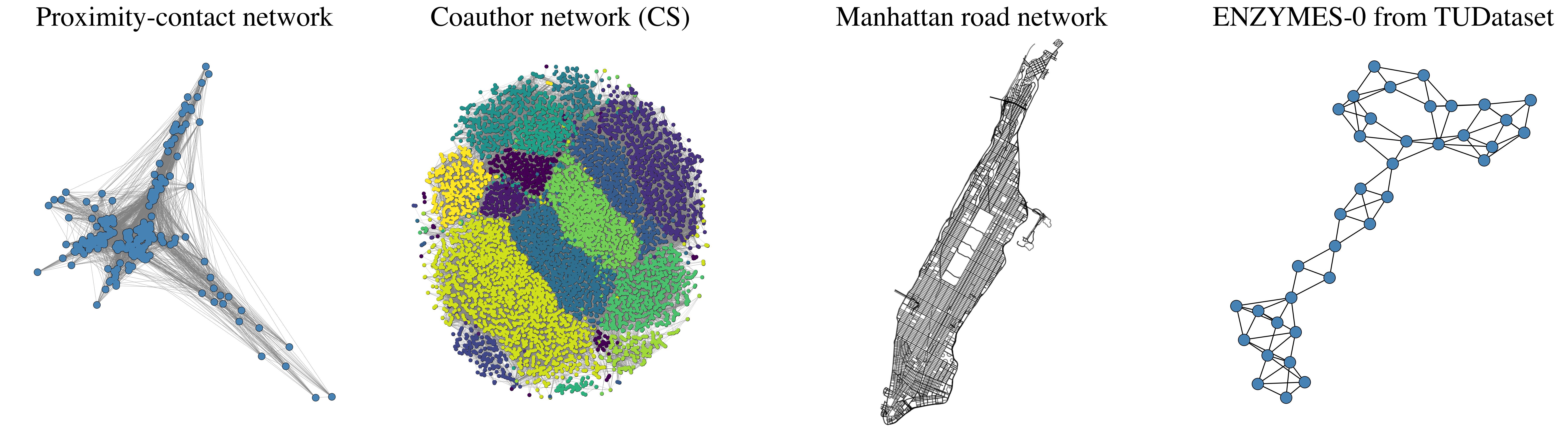}}
        \caption{Visualization of the real-world networks used in our experiments.} \label{fig realworld graph visual}
    \end{center}
    \vspace{-0.5 cm}
\end{figure}

\paragraph{Experimental setups.}
This section provides the details of our synthetic and real-world experimental settings, where we summarize the statistics of the underlying graphs and functions in Table \ref{tab statistics}, and then visualize them in Figure~\ref{fig synthetic graph visual} for synthetic problems and Figure~\ref{fig realworld graph visual} for real-world problems. 

\begin{table}[ht!]
    \centering
    \caption{Summary statistics of the underlying graphs used in our experiments.} \label{tab statistics}
    \renewcommand{\arraystretch}{1.25}
    \setlength{\tabcolsep}{6 pt} 
    \resizebox{0.99\textwidth}{!}{
    \begin{tabular}{l p{4cm} c p{4.3cm} c c}
        \toprule
        \textbf{Type} & \centering \textbf{Underlying Function} & \textbf{Search Space} & \centering \textbf{Underlying Graph} & \textbf{\# Nodes} & \textbf{\# Edges} \\
        \midrule
        \multirow{4}{*}{\textbf{Synthetic}} & Avg. Eigenvector centrality & Node subsets & BA $(m=5)$ & $10k$ & $50k$ \\
        & Avg. Degree centrality & Node subsets & WS $(k=10, p=0.1)$ & $1k$ & $5k$ \\
        & Avg. PageRank scores & Node subsets & SBM $(\text{\# clusters}=4)$ & $1k$& $7k$\\
        & Avg. Ackley function & Node subsets & 2D-GRID & $5k$ & $10k$ \\
        \midrule
        \multirow{5}{*}{\textbf{Real-world}} & Population infection time & Node subsets & Proximity contact network & $223$ & $6k$ \\
        & Individual infection time & Node subsets & SBM $(\text{\# clusters}=4)$ & $1k$ & $7k$ \\
        & Expected \# nodes influenced & Node subsets & Coauthor network (CS) & $18k$ & $163k$ \\
        & Network transitivity & Edge subsets & Manhattan road networks & $5k$ & $8k$ \\
        & Change of GNN predictions & Edge subsets & Molecule graphs (ENZYMES) & $37$ & $84$ \\ 
        \bottomrule
    \end{tabular}}
\end{table}

We closely follow the standard setups in BO literature \cite{baptista2018bayesian,eriksson2019scalable,hvarfner2022pibo} and investigate subset sizes of $k=[4,8,16,32]$, where we use $Q=4000$, $\texttt{failtol}=30$ and set $\texttt{restart\_method}$ to the best-queried combo-node for all experiments. Specifically, we query 300 times and repeat 20 times with different random seeds for each task, in which the mean and standard error of the cumulative optima are reported for all methods. For simplicity, we use a diffusion kernel~\cite{oh2019combinatorial} with automatic relevance determination and adopt \textit{Expected Improvement}~\cite{jones1998efficient} as the acquisition function.

Similar to standard BO setups, we initialize our algorithm with 10 queries by simple random walks on the original graph when $k \geq 16$, except for the influence maximization experiment where we use 30 initial queries at $k=32$ since the underlying graph is relatively large with $|\mathcal{V}|\approx 18k$. In addition, we also use simple random initialization of 30 queries for Ackley on 2D-grid and road resilience testing experiments; and 10 queries for GNN attack on molecules, since their underlying graphs contain relatively weak structural information, which are not suitable for graph-related initialization methods such as random walk. After evaluating the initial query locations, we select the best query as the starting point for all baselines to ensure a fair comparison.

\paragraph{Hardware and running time.} All the experiments are conducted on a computing cluster of 96 \texttt{Intel-Xeon@2.30GHz} CPU cores with 250 GB working memory. The running times for synthetic problems are typically under 1 hour with parallel computing. For real-world problems, except for the simulation-based experiments that take around 12 hours due to the evaluation of large numbers of simulations, the rest experiments will normally be finished within 1 hour with parallel computing.

\paragraph{Baselines.} We consider the following baselines in our experiments.
\begin{itemize}[leftmargin=20pt]
    \item \textbf{Random Search} which randomly samples $k$ nodes from the original graph at each iteration and performs well compared to other graph-based methods if the underlying function is less smooth on the combinatorial graph or less correlated with the original graph structure. 
    \item  \textbf{k-Random Walk} which maintains $k$ independent random walks on the original graph that forms a subset of $k$ nodes at each step, featuring a fast-exploration characteristic from the starting nodes, and works particularly well for exploration-heavy tasks.
    \item  \textbf{k-Local Search} takes a similar approach with the \texttt{k-Random Walk} baseline on the original graph. However, it will only proceed to the next neighbors if the current $k$-node subset is a better query compared to the previous ones, otherwise, it will hold at the same nodes and re-execute the random walk until a better location is found.    
    \item \textbf{BFS and DFS} which explore function values on graphs by starting with an initial node and then traveling the graph according to depth or breadth. Specifically, BFS exploits all nodes at the current depth before moving to the next depth, while DFS explores nodes as far as possible along each branch before backtracking. Both methods operate on the combo-graph.
    \item \textbf{Local Search} which also travels on the the proposed combo-graph by randomly selecting a node from the neighbors of the best-visited node at each iteration. If all neighbors of the best node have been queried, the algorithm will then restart from new a random location. Notably, this method can also be viewed as a BO using a random surrogate with no acquisition function and hence serves as a good indicator for BO's behavior.
\end{itemize}

\subsection{Synthetic Experiments}\label{app sec synthetic}
\begin{figure}[ht]
    \begin{center}
        \centerline{\includegraphics[width=\textwidth]{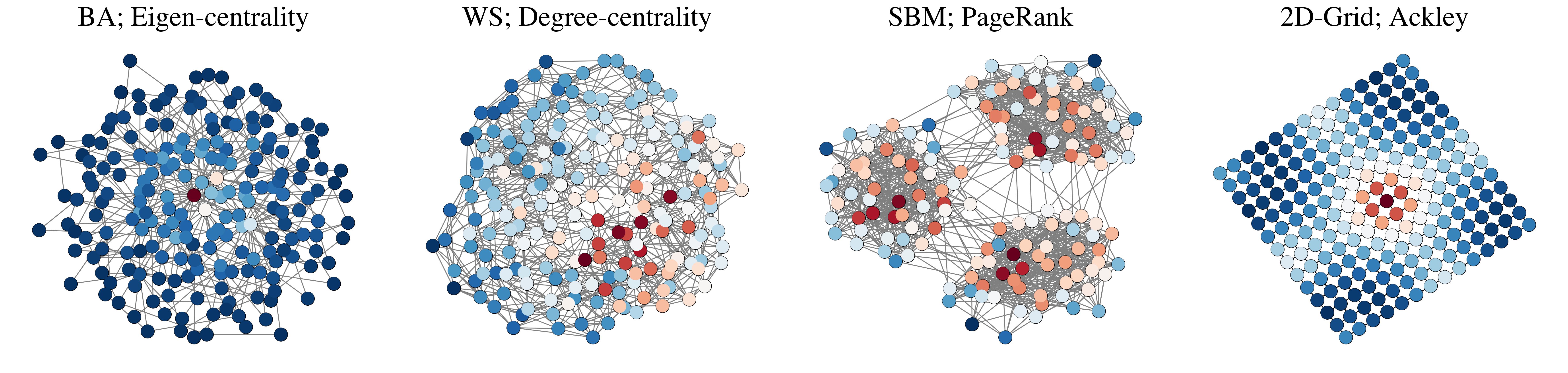}} 
        \vspace{-0.2 cm}
        \caption{Visualization of random graphs and underlying functions used in our synthetic experiments. Specifically, node color represents \textit{eigenvector centrality} on BA network, \textit{degree centrality} on WS network, \textit{PageRank} on SBM network, and \textit{Ackley} function on 2D-Grid. Note that under the synthetic settings, we take the \textbf{average} over the $k$ elements within a node subset as the underlying function. } \label{fig synthetic graph visual}
    \end{center}
    \vspace{-0.5 cm}
\end{figure}

In this section, we introduce the random graphs and synthetic functions used in our synthetic experiments, where a visualization can be found in Figure~\ref{fig synthetic graph visual}.
\paragraph{Random graphs used in experiments.}
We consider the Barab\'{a}si-Albert (BA)~\cite{barabasi1999emergence} network, Watts-Strogatz (WS)~\cite{watts1998collective} network, stochastic block model (SBM)~\cite{holland1983stochastic}, and 2D-grid as the underlying graphs in our synthetic experiments, which are explained in the following part.

\begin{itemize}[leftmargin=20pt]
    \item \textbf{Barabási-Albert network} is constructed using a preferential attachment mechanism. Specifically, we start with $m_0$ initial nodes and then gradually add new nodes one at a time, where each new node $v_i$ is connected to $m$ existing nodes with a probability $P(v_i)$ proportional to the number of links that the existing nodes already have, which can be mathematically expressed as:
    \[ P(k_i) = \frac{k_i}{\sum_{j} k_j}, \]
    where $k_i$ is the degree of node $v_i$ and the sum is over all pre-existing nodes.

    \item \textbf{Watts-Strogatz network} explains the "small-world" phenomena in a variety of networks by interpolating between a regular lattice and a random graph. In particular, the model first starts with a regular ring lattice of $N$ nodes where each node is connected to $K$ nearest neighbors (i.e. $NK/2$ total edges), then rewires $K/2$ edges for each node with probability $p \in [0, 1]$ to a random node in the network while avoiding self-loops and duplicate edges.

    \item \textbf{Stochastic Block Model} divides $N$ nodes into $K$ communities where each community $i$ has a predetermined size $N_i$. Then, the probability of an edge between nodes in cluster $i$ and $j$ is defined by a matrix $\bf P$ of size $K \times K$, where ${\bf P}_{ij}$ represents the probability of an edge between nodes in cluster $i$ and cluster $j$. The adjacency matrix $\bf A$ of the network is then generated from $\bf P$ such that entry ${\bf A}_{uv}$ for an arbitrary node-pair $u$ and $v$ is a Bernoulli random variable:
    \[ {\bf A}_{uv} \sim \text{Bernoulli}({\bf P}_{c_u c_v}), \]
    where $c_u$ and $c_v$ are the cluster memberships of nodes $u$ and $v$, respectively. In our experiment, for simplicity, we consider an SBM of $1k$ nodes with $K=4$ clusters in equal size, and fix the inter-cluster probability $p_{in}=5\times 10^{-2}$ and intra-cluster probability $p_{out} = 10^{-3}$.
\end{itemize}

\paragraph{Synthetic underlying functions used in experiments.}
We consider eigenvector centrality, degree centrality, PageRank scores, and Ackley function as the (base) underlying function on the aforementioned random graphs. Note that we first use these base functions to assign a scalar value to each node in the graph, and then take the average within the subset as the final underlying function. Such synthetic setup will enable us to track the difference between our current best query and the ground truth, which is denoted as \texttt{Regret} (to minimize) when we present the results. 

\begin{itemize}[leftmargin=20pt]
    \item \textbf{Eigenvector centrality} is a measure of the influence of a particular node $v_i$ in the network, which is defined as the solution $\bf x$ to the following equation:
    \[ {\bf A} \mathbf{x} = \lambda \mathbf{x}, \]
    where $\lambda$ is the largest eigenvalue of the adjacency matrix $\bf A$. Note that the solution $\bf x$ is also the eigenvector corresponding to $\lambda$, and hence the name eigenvector centrality.

    \item \textbf{Degree centrality} measures the number of links incident to a particular node $v_i$, which, for an undirected network of $N$ nodes, can be expressed as:
    \[ DC(v_i) = \frac{\deg(v_i)}{N-1}, \]
    where $\deg(v_i)$ is the degree of node $v_i$. The underlying intuition is to normalize the degree of each node by the maximum possible degree $N - 1$ in the network. 

    \item \textbf{PageRank} is a variant of eigenvector centrality originally developed by Google~\cite{brin1998anatomy} to rank web pages according to their ``importance'' scores. It incorporates random walks with a damping factor and can be mathematically formulated in the following form:
    \[PR(v_i) = \frac{1-d}{N} + d \sum_{v_j \in \mathcal{N}(v_i)} \frac{PR(v_j)}{\deg^{\text{out}}(v_j)}, \]
    where $d$ is the damping factor (typically set to 0.85), $N$ is the total number of nodes, $\mathcal{N}(v_i)$ denotes the in-neighbors of node $v_i$, and $\deg^{\text{out}}(v_j)$ is the out-degree of node $v_j$. 

    \item \textbf{Ackley function} is a non-convex function with multiple local optima and has been widely used for testing optimization algorithms, which can be expressed in the following 2-D form: 
    \[ f(x, y)=-20 \exp \left(-0.2 \sqrt{0.5\left(x^2+y^2\right)}\right)-\exp (-0.5(\cos 2 \pi x+\cos 2 \pi y))+20+\exp (1). \]
    Additionally, a random Gaussian noise $\sigma=0.5$ is added to the original function to alter the smoothness property of the graph signal defined over the 2D-grid:
    \[ \hat{f}(x, y) = f(x, y) + \epsilon, \text{ with } \epsilon \sim N (0, \sigma^2).\]

    Note that as the underlying graph (2D-grid) in this experiment contains little structural information compared to other graphs such as BA and WS, instead of using the random walk initialization, we adopt a simple random initialization method of 30 queries before searching and use the best query as the starting location for all methods. 
    
\end{itemize}

\subsection{Flattening the curve in Epidemics}\label{app SIR}
In this experiment, our goal is to protect an optimal subset of $k$ nodes (individuals) in a contact network to maximally slow down an epidemic process simulated by a diffusion model SIR \cite{kermack1927contribution}. The contact network \cite{stehle2011high} used in this experiment is collected by proximity sensors from a primary school in France, which contains $236$ nodes and $5,899$ edges.

\paragraph{The \textit{Suspicious}, \textit{Infected}, \textit{Recovered} simulation model.} In the SIR model based on a network $\mathcal{G}=\{\mathcal{V}, \mathcal{E}\}$, each node has three statuses: \textit{Suspicious}, \textit{Infected} and \textit{Recovered}. Starting with a fraction of $p$ initial infectious nodes in the population, at time step $t \in \{1, ..., T\}$, each infected node has a probability $\beta$ to infect another node if they are neighbors, and meanwhile has a probability $\gamma$ to transit to Recovered status. Once a node is at Recovered status, it can not be infected again (e.g. quarantined, immune, or vaccinated). More formally, let ${\bf x}_{v,t} \in \{I, S, R\}$ denote the node status (Infected, Susceptible, Recovered) and $\mathcal{S}_{I,t}, \mathcal{S}_{S,t}, \mathcal{S}_{R,t}$ denote the set of nodes in each category at time $t$, the model can be mathematically formulated as:

\begin{equation}
    \forall v \in \mathcal{S}_{I, t}, \left\{\begin{array}{l}\mathbb{P}\left[\mathbf{x}_{v, t+1}=R\right]=\gamma \\ \mathbb{P}\left[\mathbf{x}_{v, t+1}=I\right]=1-\gamma\end{array}\right.
\end{equation}
\begin{equation}
    \forall v \in \mathcal{S}_{S, t},\left\{\begin{array}{l}\mathbb{P}\left[\mathbf{x}_{v, t+1}=I\right]=1-(1-\epsilon) \times(1-\beta)^{\left|N(v) \cap \mathcal{S}_{I, t}\right|} \\ \mathbb{P}\left[\mathbf{x}_{v, t+1}=S\right]=(1-\epsilon) \times(1-\beta)^{\mid N(v) \cap \mathcal{S}_{I, t}}\end{array}\right.
\end{equation}
\begin{equation}
    \forall v \in \mathcal{S}_{R, t}, \mathbb{P}\left[\mathbf{x}_{v, t+1}=R\right]=1
\end{equation}

where $\epsilon \in [0,1]$ is a parameter representing a probability of spontaneous infection from unknown factors~\cite{wan2023bayesian}. Since the above model implies a random simulation process, a large number of Monte Carlo samples is typically required when estimating the expectation of certain functions based on SIR, which is often expensive to evaluate and optimize.

\begin{figure*}[ht]
    \begin{center}
        \centerline{\includegraphics[width=\textwidth]{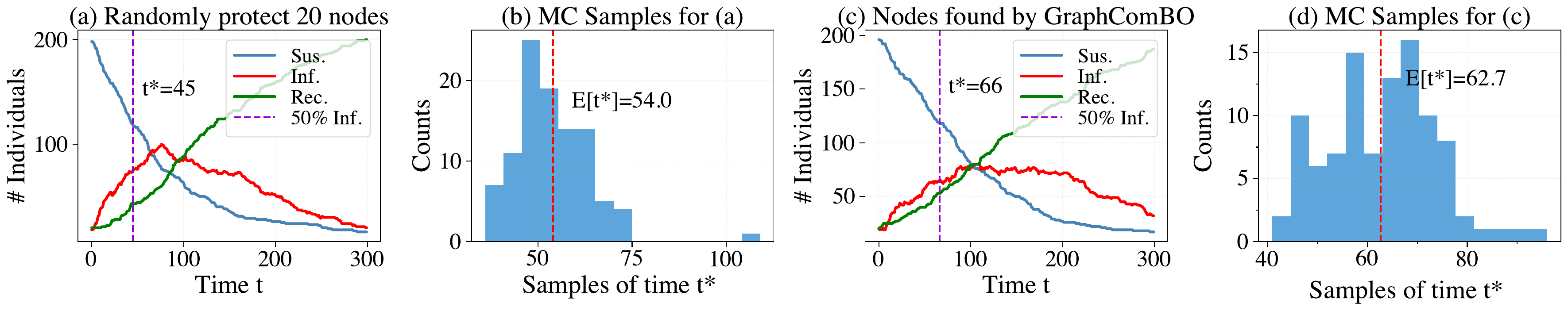}} \vskip -0.2 cm
        \caption{Demonstration of SIR and its simulations on the real-world proximity contact network.} \label{fig SIR demo}
    \end{center}
    \vskip -0.3 cm
\end{figure*}

\paragraph{Flattening the curve with SIR.}
As healthcare resource is often limited (e.g. hospitals, vaccines, quarantine centers), one is usually interested in slowing down the transmission speed of the epidemic process by protecting the most ``important'' individuals, which prevents the public health system from breaking down due to the sudden shortage of its capacity~\cite{maier2020effective}. We demonstrate this idea with SIR in the above contact network with $N=100$ simulations in Figure \ref{fig SIR demo}, in which each run has an initial fraction of $p=0.1$ population infected, an infection rate of $\beta=10^{-3}$, a recovery rate of $\gamma=10^{-2}$, and no spontaneous infection $\epsilon=0$. Note that the same settings have been used in the main experiments. 

To protect nodes from infection, we set the chosen $k$-node subset to the Recovered status at the beginning of each simulation, then record the time $t^*$ that 50\% population is infected. After obtaining the results from $N=100$ simulations, we record the mean infection time $\mathbb{E}[t^*]$ as our underlying function, which we aim to maximize (i.e. delay the time when reaching half-population infection). From Figure~\ref{fig SIR demo}, we can observe a clear curve-flattening effect when protecting 20 nodes selected by BO (plots c \& d) compared to randomly choosing 20 nodes (plots a \& b) in the network. Note that to make the surrogate fitting more numerically stable, we also map $\mathbb{E}[t^*]$ to the range $[0, 1]$ by dividing a constant of the maximal number of iteration $T=120$.

\subsection{Tracing Patient-zero in the Community}\label{app patient zero}
\begin{figure}[t]
    \begin{center}
        \centerline{\includegraphics[width=\textwidth]{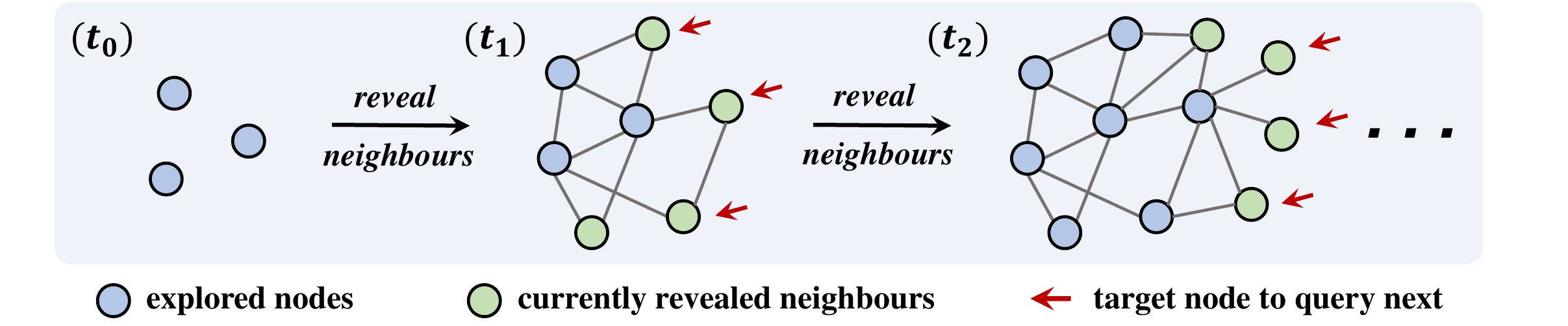}} \vspace{-0.1 cm}
        \caption{Demonstration of the patient-zero tracing setting, where the underlying contact network is not fully observed initially. To gradually reveal the graph structure, at each step $t$, we query $k$ nodes (i.e. record the first times they are infected) and reveal their neighbors (e.g. interview the patients and obtain their contacts). The objective is to find the $k$ patients of the earliest infection time.} \label{fig patient zero demo}
    \end{center}
    \vskip -0.5 cm
\end{figure}

In disease transmission analysis such as AIDS and COVID-19, one would be interested in identifying the earliest individuals (patient-zero) infected in the community \cite{mckay2017patient, wan2023bayesian}. Nevertheless, such a process is often time-consuming since it requires interviewing patients to obtain their infection dates and then gradually revealing the transmission network by interviewing their close contacts (i.e., the graph is not known a prior), as illustrated in Figure~\ref{fig patient zero demo}.

\paragraph{Individual infection time.} In this task, we use the aforementioned SIR model and SBM network to simulate a disease contagion across multiple communities, where the goal is to identify the earliest $k$ individuals infected in the whole network. Specifically, at a certain time step $T$ during the epidemic (set to $T=100$ in the experiment), for every node in $\mathcal{S}_{I,T} \cup \mathcal{S}_{R,T}$ we denote $\tau_v$ as the time of infection for node $v$ and map it into a scalar value $f(v) \in [0, 1]$ via the following transformation: 
\begin{equation}
    \forall v \in \mathcal{V}, f(v)= \begin{cases}0 & \text { if } v \in \mathcal{S}_{S, T} \\ \left(1-\frac{\tau_v}{T}\right)^2 & \text { if } v \in \mathcal{S}_{I, T} \cup \mathcal{S}_{R, T}\end{cases}
\end{equation}
Then, we take the average within the $k$-node subset as the final underlying function in this experiment, which is maximized when the subset corresponds to the earliest $k$ nodes of infection. 

Note that in this experiment only a single SIR simulation is required, and we run 20 times with different random seeds to report the results. For reference, the parameters used in each run are: initial fraction of infected population $p=0.5\%$, infection rate \& recovery rate of $\beta=\gamma=10^{-2}$, spontaneous infection rate of $\epsilon=0.5\%$, and simulation time step of $T=100$.

\subsection{Influence Maximization on Social Networks}\label{app IC}
The \textit{Influence Maximization} (IM) problem over social networks has been widely applied to marketing and recommendations \cite{li2018influence}, where the goal is to select the optimal $k$ nodes as the seeds (i.e. source of influence) that maximize the expected number of final influenced individuals, which is typically estimated by \textit{Independent Cascading} (IC) simulation~\cite{kempe2003maximizing} and its variants. In this experiment, we consider the vanilla IC simulations on an academic collaboration network (Coauthor CS~\cite{shchur2018pitfalls}) of $18,333$ nodes and $163,788$ edges, with detailed settings discussed in the following.
\begin{figure*}[ht]
    \begin{center}
        \centerline{\includegraphics[width=\textwidth]{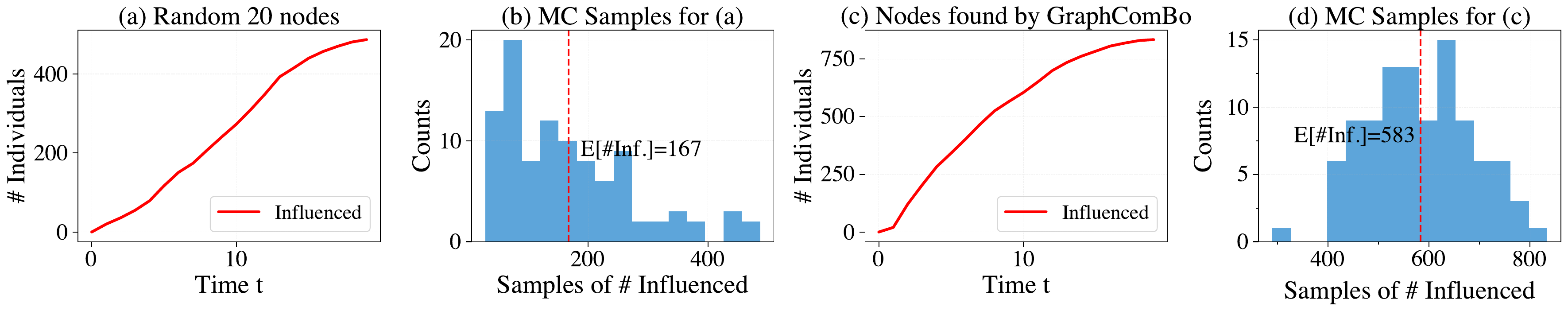}} \vskip -0.2cm
        \caption{Demonstration of Independent cascading with simulations on the CS coauthor network.} \label{fig IC demo}
    \end{center}
    \vskip -0.5 cm
\end{figure*}

\paragraph{Independent cascading.} 
The IC model takes a similar setup to the SIR model above, except that each node only takes one of the two statuses: \textit{Activated} and \textit{Inactivated}. Starting with an initial subset of $k$ activated nodes, at each time step $t \in \{1, ..., T\}$, the activated nodes will have a one-shot probability of $p$ to activate their neighbors. Once a node is activated, it will remain activated until the end of the process when there is no more new node to be activated. Likewise, calculating the expectation of functions based on IC also requires a large number of Monte Carlo simulations, which is hence computationally expensive in most cases.

\paragraph{Influence maximization with IC.}  The task of IM is to select the optimal set of $k$ nodes as the seeds of influence, such that the expected number of activated nodes at the end of IC is maximized. In our experiment, we set the activation rate to $p=0.05$ and used $N=1000$ Monte Carlo samples to estimate the expected number of final influenced individuals. In particular, the underlying function is set to be $\#\text{Activated}/|\mathcal{V}|$, that is, the fraction of activated nodes in the network. Figure \ref{fig IC demo} demonstrates the simulation results from two different strategies of selecting $k=20$ initial seeds: random selection (plots a \& b) and GraphComBO (plots c \& d), and we can observe a clear difference in their expected numbers of final activated nodes.

\subsection{Road Resilience Testing on Transportation Networks} \label{app road}
The resilience of an infrastructure network denotes its ability to maintain normal operations when facing certain disruptions \cite{liu2020review}, and one is often interested in identifying and protecting the most important nodes or edges to avoid catastrophic failure. In this experiment, we investigate the resilience of Manhattan road networks in New York City using \texttt{OSMnx}~\cite{boeing2017osmnx}, a Python tool built on \texttt{OpenStreetMap}~\cite{haklay2008openstreetmap}. Figure~\ref{fig realworld graph visual} provides a visualization of the network, where each node denotes an intersection and each edge represents a road (with network\_type set to ``drive'')

The objective here is to identify the $k$ most vulnerable roads, such that their removal will lead to the maximal drop in a certain utility function, which could be difficult to evaluate in practice if it involves simulations or real-world queries. For experimental purposes, we use network transitivity (global clustering coefficient) as a proxy estimation for this utility function (to minimize), which measures the global connectivity of the graph based on the number of triangles:
\begin{equation}
    \text{Transitivity}(\mathcal{G}) = \frac{\# \text{Triangles}}{\#\text{Triads}},
\end{equation}
where \texttt{Triad} means two edges with a shared vertex. Similar to the Ackley on 2D-grid experiment, as the underlying graph is a grid-like road network, we utilize simple random initialization of 30 queries before searching and use the best query as the starting location for all baselines. 

\paragraph{Line-graph for functions of edge subsets.}
Since the underlying function here is defined on edge (road) subsets,  we will first change the original graph $\mathcal{G}$ into its line graph $\mathcal{G}_{line}$, where each node on the line graph $\mathcal{G}_{line}$ represents an edge in $\mathcal{G}$, and two nodes on $\mathcal{G}_{line}$ are adjacent if and only if their corresponding edges in $\mathcal{G}$ share a common endpoint. Then, the line graph will be used as the underlying graph in our proposed framework. Note that this procedure is independent of the underlying function evaluation, which is still on the original graph with certain black-box processes.

\subsection{Black-box Attacks on Graph Neural Networks} \label{app gnn}
In this task, we conduct adversarial attacks on graph neural networks (GNN) under a challenging black-box setting \cite{sun2022adversarial}, where the attacker has no access to model parameters but only a limited number of queries for the outputs. Considering a GNN pre-trained for graph-level classification tasks, for a particular target graph under attack, our goal is to mask $k$ edges on this input graph, such that the output from GNN (at softmax) will be maximally perturbed from the original output.

\paragraph{Perturbation via edge-masking.} Concretely, we use GIN~\cite{xu2019powerful} as the victim GNN and pre-train it on the TUDataset ENZYMES~\cite{morris2020tudataset} for small molecule classification, and measure the change in GNN prediction after perturbation by the Wasserstein distance, which can be formulated as follows:
\begin{equation}
    f(\mathcal{S}) = W_1 \Big( g\big(\mathcal{G}\big), \; g\big(\Phi(\mathcal{G},\mathcal{S})\big) \Big),
\end{equation}
where $f(\mathcal{S})$ is the underlying function defined on the $k$-edge subset $\mathcal{S}$ (to maximize), $W_1$ is the empirical Wasserstein-1 distance, $g$ denotes the pre-trained GNN at softmax, and $\Phi(\mathcal{G},\mathcal{S})$ is a perturbation on $\mathcal{G}$ by masking the $k$-edge subset $\mathcal{S}$, which will lead to a perturbed graph $\mathcal{G}'$.

For reference, the GIN model has $3$ hidden layers with $64$ hidden dimension and is trained for $200$ epochs by Adam~\cite{kingma2014adam} with a learning rate of $10^{-3}$. The pre-trained model achieves $99.5\%$ accuracy on the dataset, and we use the first graph (index=0) from the dataset as the target graph under attack. Note that no training/testing split is needed here as we are conducting attacks on pre-trained GNN.

Since the underlying function is defined on edge subsets of small molecule graphs, we will use the above line graph operation again and only search for $100$ queries, where simple random initialization is also adopted for $10$ queries before searching.

%% file: 9Complexity.tex
\section{Algorithm Complexity}\label{app complexity}
Overall, the computational complexity of the proposed method for optimizing a function defined on $k$ nodes in a graph of size $N$ mainly comes from (1) constructing a combo-subgraph of size $Q$ and (2) fitting the surrogate model.

\begin{enumerate}[leftmargin=20pt]
    \item Suppose at recursion $\ell \in \{1, 2, 3, ...\}$ we found $M_{\ell}$ new combo-nodes ($M_{0}$ = 1). We will first need $kM_{\ell-1}$ dictionary lookup operations to find the neighbors of nodes in the $k$-node subset for each $M_{\ell-1}$ input combo-nodes, and then $M_{\ell}$ concatenations to create $M_{\ell}$ new combo-nodes. The recursion repeats until $\sum_{\ell=0}^L M_{\ell} \geq Q$ at $\ell = L$, which leads to a total of $k\sum_{\ell=0}^{L-1} M_{\ell} \leq kQ$ dictionary lookups at $\mathcal{O}(kQ)$, plus around $Q$ concatenations at $\mathcal{O}(Q)$ (the last recursion $L$ will break before finish when we reach $Q$).
    
    \item The computational cost consists of two parts: (a) the Graph Fourier Transformation (GFT) and (b) computing the predictive posterior in Gaussian Processes (GP) using the Gaussian conditioning rule.
    \begin{enumerate}
        \item The GFT requires eigendecomposing the graph Laplacian matrix: $\bf L=U \Lambda U^\top$, which typically costs $\mathcal{O}(N^3)$ for a graph of $N$ nodes.
        \item To obtain the predictive posterior from a GP by Gaussian conditioning rules (explained in section 2 line 105), we need to compute the inverse of the kernel matrix $K^{-1}_{1:t}$ for the observed $t$ datapoints, which requires $\mathcal{O}(t^3)$. Since $t <= N$, the maximum complexity for this term is also at $\mathcal{O}(N^3)$ for a graph of $N$ nodes.
    \end{enumerate}
\end{enumerate}

However, since the surrogate model operates on a subgraph of size $Q$, we can limit the computational cost to $\mathcal{O}(Q^3)$ and only need to re-construct the combo-subgraph and re-compute its eigenbasis when the center changes (i.e. when finding a better query location). Note that the computational cost from (a) is at $\mathcal{O}(Qk)$ which is insignificant compared to $\mathcal{O}(Q^3)$ in practice. This also leads to an efficient memory consumption where we only need to store a combo-subgraph of size $Q$ with its cached eigenbasis (a $Q\times Q$ matrix) during the search.

\section{Proofs of Lemmas in \S\ref{sec combo-graph}} \label{app proof}
In this section, we provide proofs for \textbf{Lemma~\ref{lemma hop}} and \textbf{Lemma~\ref{lemma linear}} in \S\ref{sec combo-graph}.

\textbf{Lemma 3.2.} \textit{In the proposed combo-graph, at most $\ell$ elements in the subset will be changed between any two combo-nodes that are $\ell$-hop away.}
\begin{proof}
    Considering an arbitrary combo-node $\hat{v}_i = (v_i^{(1)}, v_i^{(2)}, ..., v_i^{(k)})$ of $k$ elements as the center of an $\ell$-hop ego combo-subgraph on the proposed combinatorial graph $\Tilde{\mathcal{G}}^{<k>}$. According to \textbf{Definition~\ref{def combo-graph}}, there will be strictly 1 element in difference between two neighboring combo-nodes, which implies that the $1^\text{st}$-hop neighbors of $\hat{v}_i$ will have one different element, the $2^\text{nd}$-hop neighbors will have one or two different element(s), the $3^\text{rd}$-hop will have one, two, or three different element(s), ..., and by induction, we can conclude that at hop-$\ell$ there will be at most $\ell$ elements in difference.
\end{proof}

\textbf{Lemma 3.3.} \textit{The degree of combo-node $\hat{v}_i$ increases linearly with $k$ and is maximized by the subset of nodes with top $k$ degrees: $\deg(\hat{v}_i) = \sum_{j=1}^k | \mathcal{N}(v_i^{(j)}) \setminus \{v_i^{(j')}\}_{j'\neq j}^k |$.}
\begin{proof}
    The degree $\deg(\hat{v}_i)$ of an arbitrary combo-node $\hat{v}_i = (v_i^{(1)}, v_i^{(2)}, ..., v_i^{(k)})$ is a linear combination over $k$ constant terms, where each term $j \in \{1,...,k\}$ equals the number of neighbors $\mathcal{N}(v_i^{(j)})$ of an element node $v_i^{(j)}$ that are not inside the combination. As the maximum term is capped by $|\mathcal{V}|-1$, which is the largest possible degree in the original graph of an arbitrary structure, we conclude that the combo-node degree will increase linearly with $k$.
\end{proof}

%% file: 8KernelDetails.tex
\section{Details of the Kernels on Graphs}\label{app kernel}
\paragraph{Common choices of kernels on graphs.} Following the discussion in \S\ref{sec GP}, we analyze the performance of four kernels on the combinatorial graph and their details are summarized in Table~\ref{tab kernels}. Specifically, we consider: \textbf{Polynomial} \cite{defferrard2016convolutional} that consists of polynomials of the eigenvalue at order $\eta \in \mathbb{Z}_{\geq 1}$, where the hyperparameters are the coefficient for each order; \textbf{Sum-of-Inverse Polynomials} \cite{wan2023bayesian} which is a variant of the polynomial kernel that takes a scaled harmonic mean of different degrees; \textbf{Diffusion} \cite{oh2019combinatorial} that penalizes the magnitude of the frequency (eigenvalue), and we also consider its implementation with the automatic relevance determination (ARD) strategy.

\begin{table}[t]
    \begin{center}  
    \caption{Summary of some common kernels on graphs.} \label{tab kernels}
    \vspace{-0.2 cm}
    \setlength{\tabcolsep}{6 pt} 
	\resizebox{0.9\textwidth}{!}{
    	\begin{tabular}{l l l}
    		\toprule
            Choice of Kernel & Regularization $r(\lambda_p)$ & Kernel $K_n(\mathcal{V}, \mathcal{V})$ \\
    		\midrule
    		Diffusion   & $\exp(\beta_p \lambda_p)$  & $\sum_{p=1}^n \exp(-\beta_p \lambda_p){\bm u}_p{\bm u}_p^\top$\\
    	    Polynomial  & $\sum_{j=1}^{\eta-1}\beta_j\lambda_p^{j} + \epsilon$  & $\sum_{p=1}^n \big( \sum_{j=1}^{\eta-1}\beta_j\lambda_p^{j} + \epsilon \big)^{-1}{\bm u}_p{\bm u}_p^\top$ \\
            \multirow{2}{3cm}{Sum-of-Inverse Polynomials} & \multirow{2}{4.3cm}{$\big(\sum_{j=1}^{\eta-1} (\beta_j\lambda_p^{j}+\epsilon)^{-1}\big)^{-1}$}  & \multirow{2}{6cm}{$\sum_{p=1}^n \big(\sum_{j=1}^{\eta-1} (\beta_j\lambda_p^{j}+\epsilon)^{-1}\big) {\bm u}_p{\bm u}_p^\top$} \\
            \\
    		\bottomrule
    	\end{tabular}
    }
    \end{center}
\end{table}

The polynomial and sum-of-inverse polynomials kernels have $\eta$ hyperparameters ${\bm \beta} = [\beta_0, \cdots, \beta_{\eta -1}]^\top$ that are constrained to be non-negative to ensure a positive semi-definite covariance matrix. Meanwhile, we maintain the settings in the previous work \cite{wan2023bayesian} that set $\eta$ to be $\min\{5, \texttt{diameter}\}$, which strikes a balance between expressiveness and regularisation. Whereas in the diffusion kernel, there are $n$ hyperparameters ${\bm \beta} = [\beta_1, \cdots, \beta_n]$ to be learned and are sometimes prone to over-fitting when $n$ is large. 

\begin{figure}[ht]
    \begin{center}
        \centerline{\includegraphics[width=\textwidth]{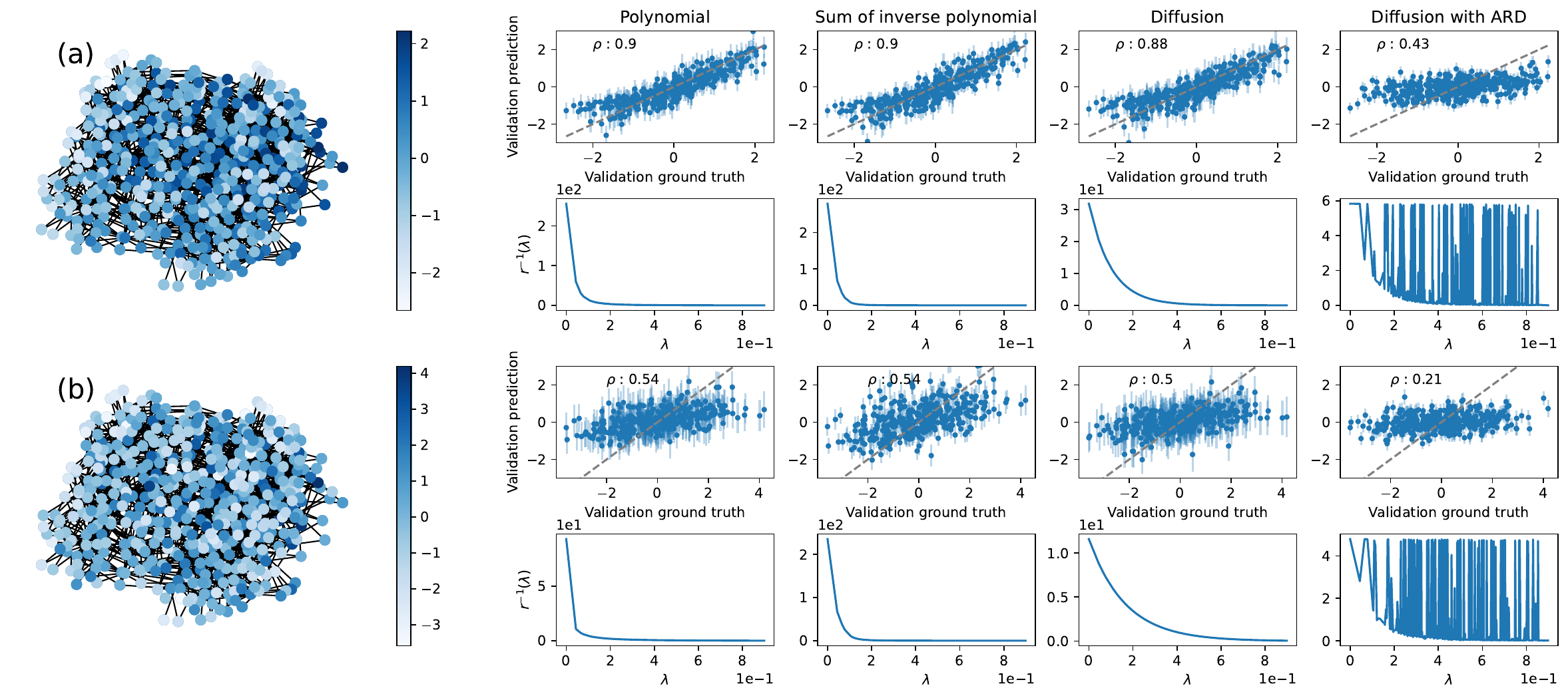}} \vskip -0.2 cm
        \caption{Kernel validation on the combinatorial graph based on a BA network ($n=20, m=2$). To design the underlying function, we take the elements from the third eigenvector and average them over $k=3$ nodes. Specifically, (a) shows the results on the testing data measured by Spearman's correlation coefficient $\rho$, and (b) shows the results when adding Gaussian noise to the ground truth.} \label{fig kernel validation ba}
    \end{center}
    \vskip -0.5 cm
\end{figure}

\begin{figure}[ht]
    \begin{center}
        \centerline{\includegraphics[width=\textwidth]{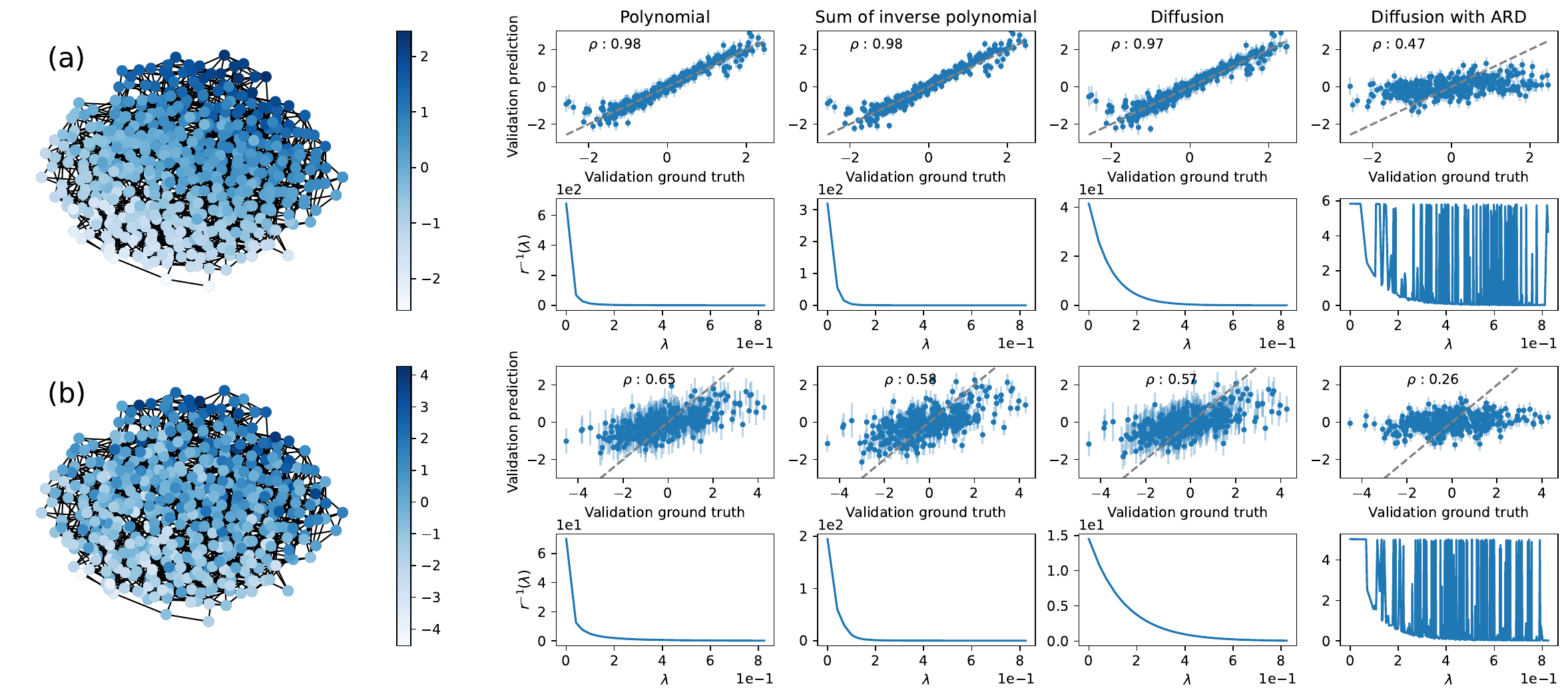}} \vskip -0.2 cm
        \caption{Kernel validation on the combo-graph based on a WS network ($n=20, k=5, p=0.2$).} \label{fig kernel validation ws}
    \end{center}
    \vskip -0.5 cm
\end{figure}

\paragraph{Kernel validation.} We consider a synthetic setting on two $20$-node networks with subset size $k=3$: a BA network ($m=2$) and a WS network $(k, p) = (5, 0.2)$, where the combinatorial graph in both networks contains $\binom{20}{3}=1140$ combo-nodes. The underlying function is designed in the following way: we first perform eigen-decomposition on the graph Laplacian matrix: $\Tilde{\bf L} = {\bf U} {\bm \Lambda} {\bf U}^\top$ where $\Tilde{\bf L} = {\bf I} - \Tilde{\bf D}^{-1/2}\Tilde{\bf A}\Tilde{\bf D}^{-1/2}$ is the normalised graph Laplacian. The eigenvalues ${\bm \Lambda}=\text{diag}(\lambda_1, \cdots, \lambda_n)$ are then sorted in ascending order and possess frequency information in the spectral domain \cite{dong2020graph}, where smaller eigenvalues indicate lower frequencies. As such, their corresponding eigenvectors ${\bf U}=[{\bm u}_1, \cdots, {\bm u}_n]$ can be used as signals of different smoothness and will be discussed in more detail in Appendix \S\ref{app smoothness}. For the current experiment, we will take the elements from the eigenvector that corresponds to the 2nd non-zero eigenvalue (which is a smooth signal on the underlying graph), and use the average over $k$ nodes as the underlying function in the combinatorial space. 

After standardization, we use 25\% combo-nodes as the training set to fit the models and validate their performance on the rest 75\% combo-nodes with Spearman's rank-based correlation coefficient $\rho$. In addition, we also consider a noisy scenario where a Gaussian noise of $\sigma=1$ is added to the original function, where their results are summarized in Figure \ref{fig kernel validation ba} for BA and Figure~\ref{fig kernel validation ws} for WS. We observe that all kernels can capture the original signal except for Diffusion with ARD, which learns a non-smooth transformation on the spectrum due to its over-parameterization. Nevertheless, we found the difference in performance is insignificant when using less-smooth underlying functions in \S\ref{app smoothness}. 

%% file: 11Behaviour.tex
\section{Kernel Performance under Different Signal Smoothness}\label{app smoothness}
Following the setups in Appendix~\S\ref{app kernel} above, we now investigate how the inherited smoothness of the underlying function influences the performance of our kernels.

\begin{figure}[ht]
    \begin{center}
        \centerline{\includegraphics[width=\textwidth]{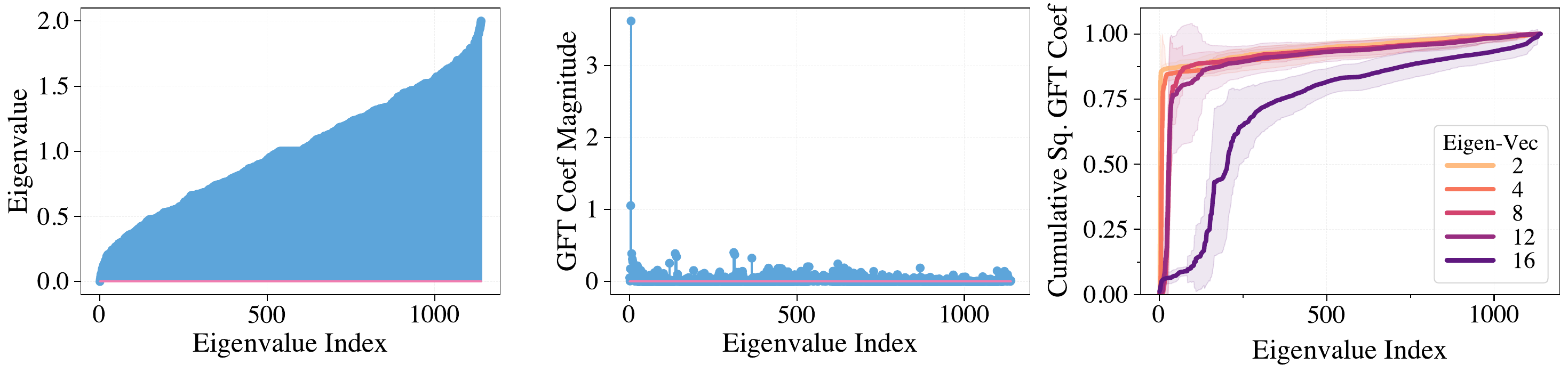}} \vskip -0.2 cm
        \caption{Smoothness of different underlying functions (average of different eigenvectors).} \label{fig smoothness j}
    \end{center}
    \vskip -0.5 cm
\end{figure}

\paragraph{The smoothness of graph signals in the combinatorial space.}
The graph Fourier transform is given by $\hat{f}({\bf \Lambda}) = {\bf U}^\top f$, which transforms the original graph signal $f$ to the frequency domain, as illustrated in the second plot from Figure~\ref{fig smoothness j}. To change the smoothness of the underlying function in the combinatorial space, we consider $j$-th eigenvector with $j \in [2,4,8,12,16]$ as the underlying signals (from the original graph) and then use the same method in \S\ref{app kernel} that takes the average over the nodes in the subset as the underlying function in the combinatorial space. To compare the smoothness among different underlying functions (in the combinatorial space), we first calculate the cumulative energy of Fourier coefficients; since we are modeling on random graphs, the process will be repeated 50 times with different random seeds, after which we plot the results on the third plot in Figure \ref{fig smoothness j} with mean and standard error. We can observe a clear trend that the underlying function in the combinatorial space becomes less smooth when using an eigenvector that corresponds to a higher frequency (larger eigenvalue). 

\begin{figure}[ht]
    \begin{center}
        \centerline{\includegraphics[width=\textwidth]{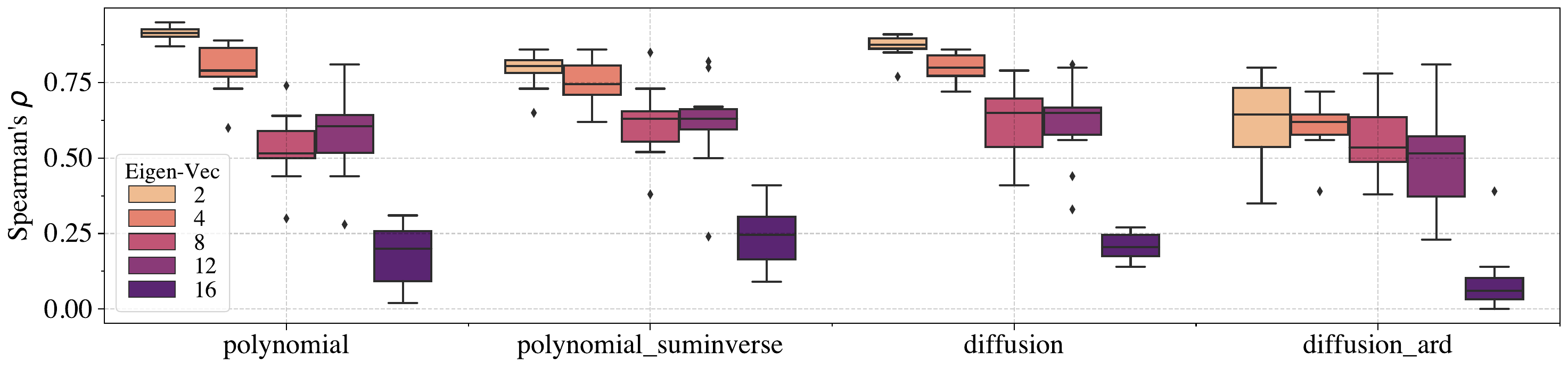}} \vskip -0.2 cm
        \caption{Performance (Spearman's rank-based correlation coefficient $\rho$) of kernels for underlying functions with different smoothness, where darker shades use eigenvectors of the higher index and thus indicate less-smooth functions.} \label{fig kernel smoothness}
    \end{center}
    \vskip -0.5 cm
\end{figure}

\paragraph{Kernel performance under different smoothness.}
With the same settings in Appendix~\S\ref{app kernel}, we validate the performance of kernels on functions of different smoothness levels in the combinatorial graph (as described above) and report their results by Spearman's correlation coefficient $\rho$ as a box-plot in Figure \ref{fig kernel smoothness}. For each kernel, we can see a clear drop in its validation performance as the function becomes less smooth, which will in turn negatively affect the performance of BO.

\section{Behavior Analysis of GraphComBO}\label{app behaviour}
\begin{figure}[ht]
    \begin{center}
        \centerline{\includegraphics[width=\textwidth]{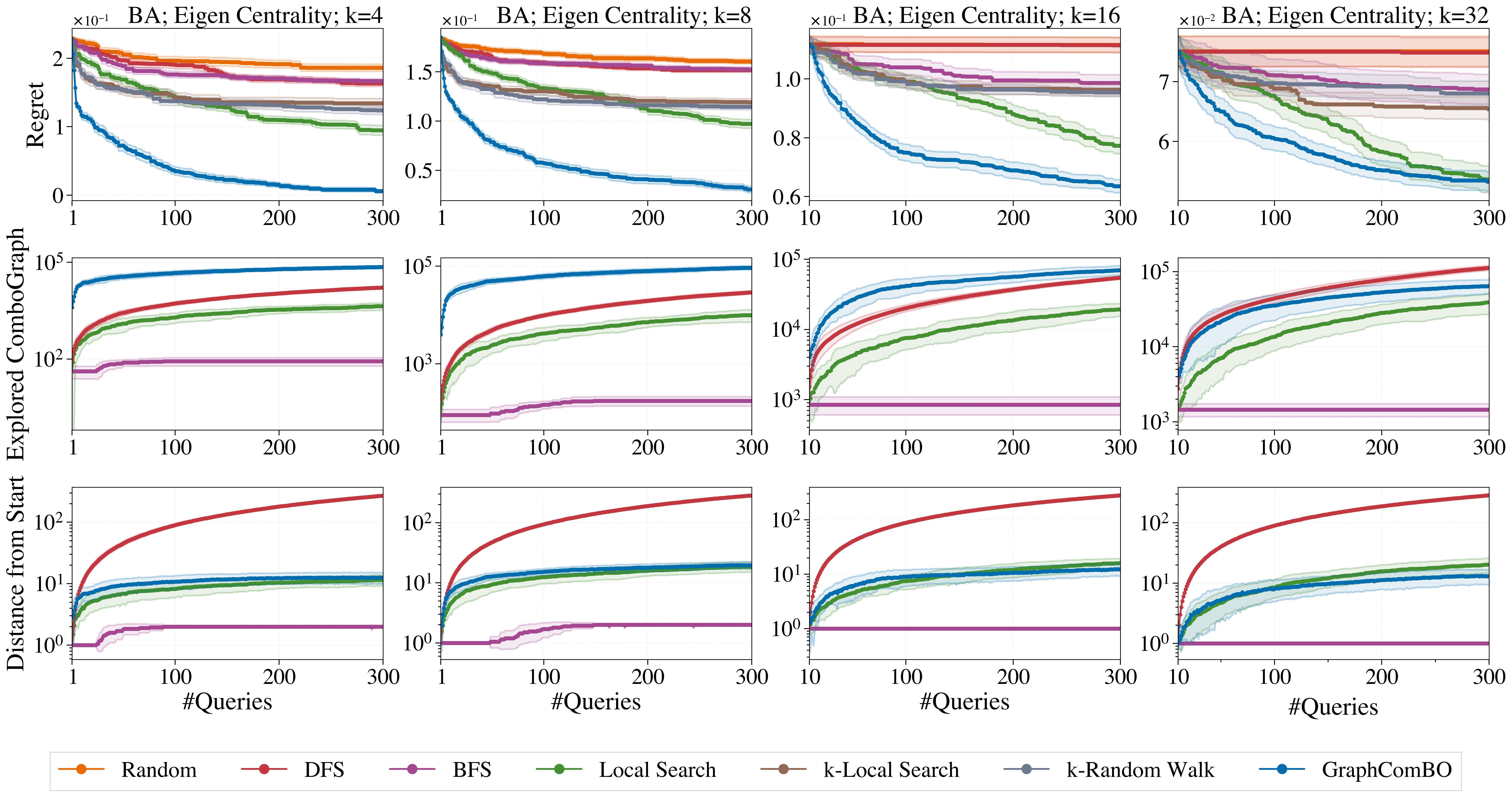}} \vskip -0.2 cm
        \caption{Behavior analysis of maximizing average eigenvector centrality on the BA network.} \label{fig behavior ba}
    \end{center}
    \vskip -0.5 cm
\end{figure}

\begin{figure}[ht]
    \begin{center}
        \centerline{\includegraphics[width=\textwidth]{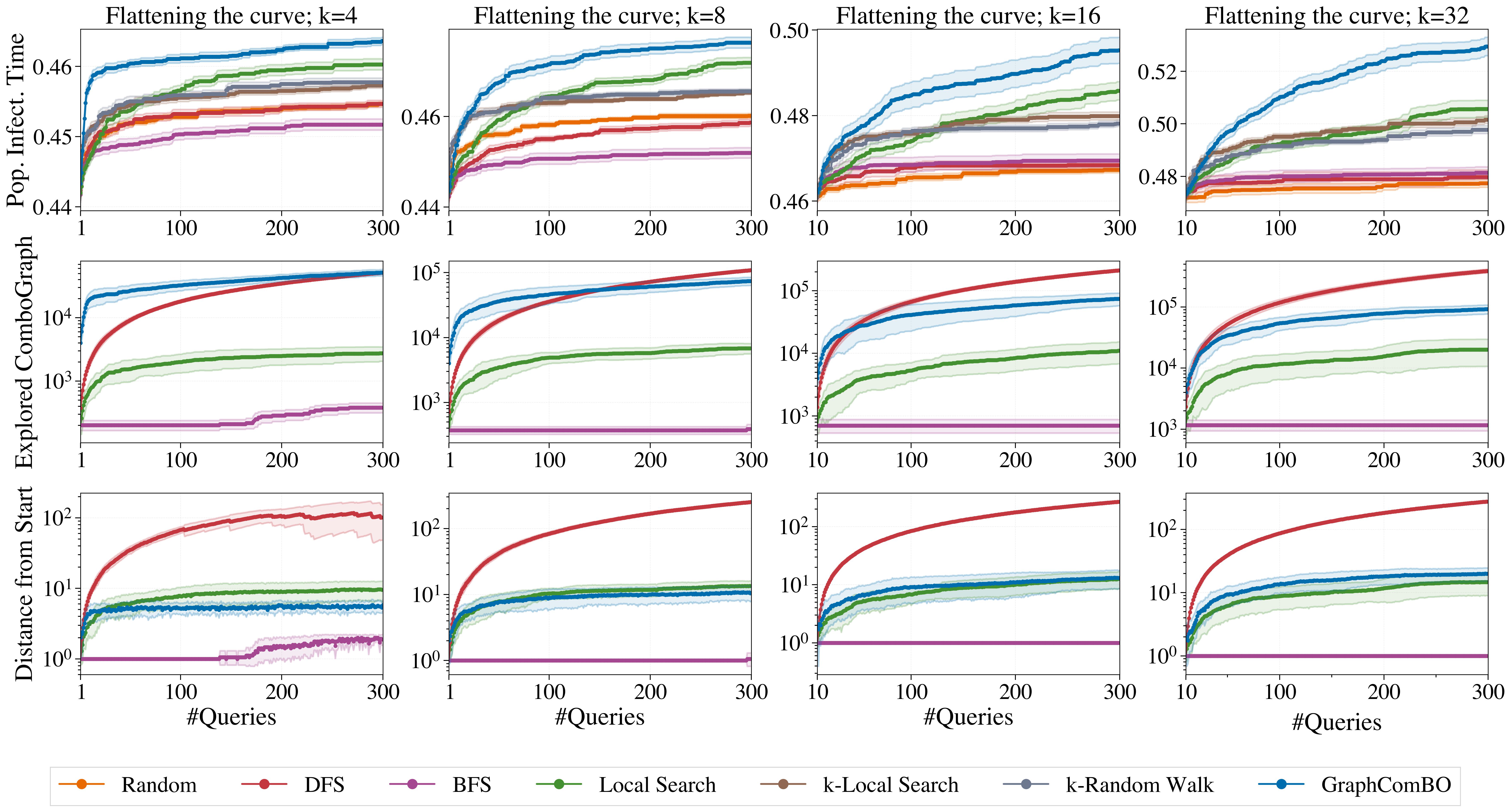}} \vskip -0.2 cm
        \caption{Behavior analysis of flattening the curve experiment on the contact network with SIR.} \label{fig behavior SIR}
    \end{center}
    \vskip -0.5 cm
\end{figure}

In this section, we provide an in-depth behavior analysis of GraphComBO from two of the main experiments: a synthetic task of maximizing the average eigenvector centrality on BA networks, and a real-world task of flattening the curve on the contact network. The results are present in Figure~\ref{fig behavior ba} and Figure~\ref{fig behavior SIR} respectively, where we also record additional information on (1) the explored combo-graph size and (2) the distance of the current combo-subgraph center to the starting location. Note that these recorders are only available for methods based on the proposed combo-graph, where all of these methods start at the same location before searching.

From both figures, it is straightforward to find that BFS and DFS behave differently given their exploration size and travel distance, in which BFS performs heavy exploitation and DFS performs large exploration. On the other hand, while BO explores more combinatorial space than local search in both experiments, we can notice the following distinctions in the source of its performance gain.

Considering the synthetic experiment results on BA network with a small $k$, we can observe that the subgraph center of GraphComBO is slightly more distant from the start location compared to the local search at the beginning, but later saturates and is caught up by local search. Such behavior also holds when $k$ increases, especially at $k=32$ where local search generally travels more distantly than GraphComBO. This implies that when $k$ is small, the performance gain may mainly come from the exploration, whereas when $k$ increases, we are losing the relative advantage of exploration and the performance gain is mainly from exploitation, which is consistent with our conjecture in \S\ref{sec subset size}. 
On the contrary, the result from flattening the curve experiment tells a different story, where GraphComBO takes a more exploitation-focused strategy compared to local search when $k$ is small, but as $k$ increases, it gradually shifts to an exploration-driven behavior.

%% file: 12COMBO.tex
\section{Comparison with COMBO} \label{app COMBO}
\begin{figure}[ht]
    \begin{center}
        \centerline{\includegraphics[width=\textwidth]{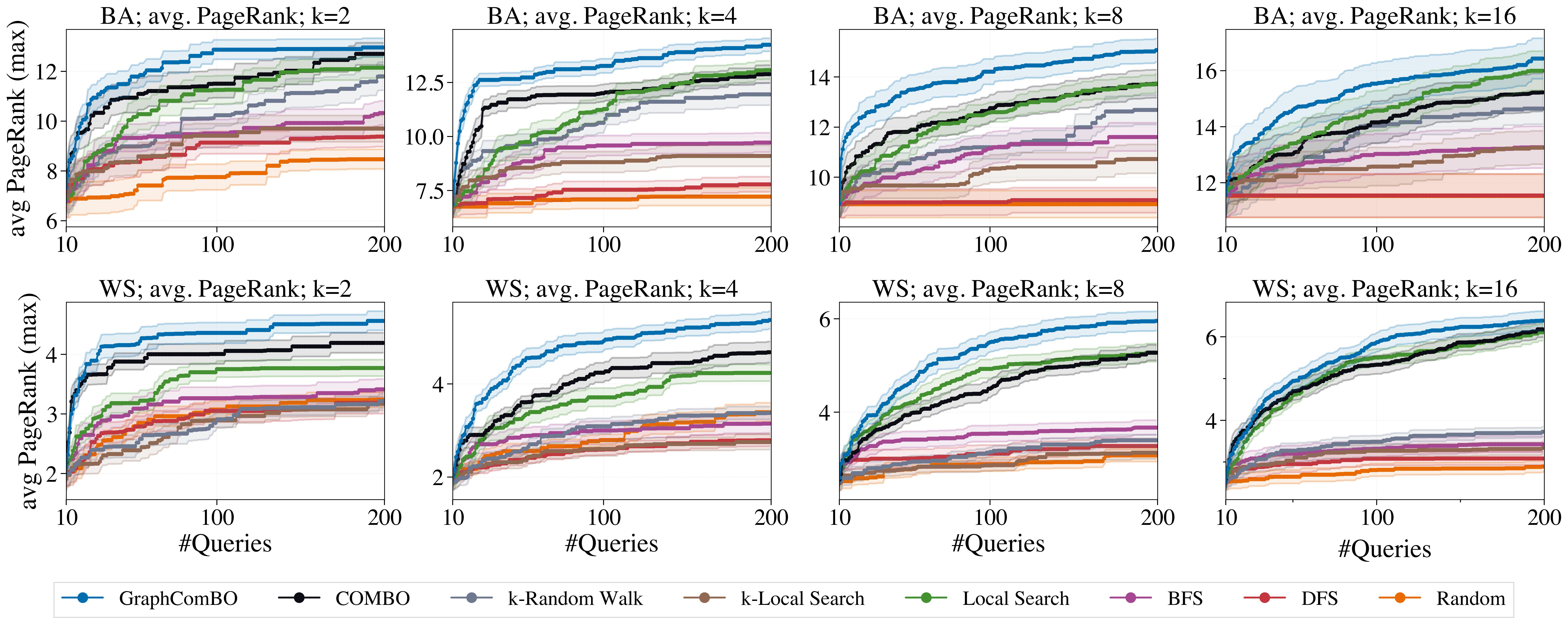}}
        \vspace{-0.2 cm}
        \caption{Comparison with COMBO in maximizing avg. PageRank on small BA and WS networks.} \label{fig combo comparison}
    \end{center}
    \vspace{-0.2 cm}
\end{figure}

In this section, we compare our method with COMBO~\cite{oh2019combinatorial} on small BA and WS graphs of $|\mathcal{V}| = 500$ (still much larger than the graphs used in COMBO's experiments), where the results in Figure~\ref{fig combo comparison} show a clear advantage of our framework over COMBO, and we make the following explanations.

To implement COMBO under our setting of $k$-node subsets from a single graph $\mathcal{G}$ of size $N$, we generate $k$ identical copies of $\mathcal{G}$ and form the $k$-node subset by drawing one node from each of the copy. This leads to a search space of $N^k$, which is the key limitation of COMBO under this setting since it is supposed to be $\binom{N}{k}$. As a result, there are many repeated and invalid locations in the search space, for example, at $k = 3$, $(1,2,3), (1,3,2), (2,1,3), ...$ are different subsets in COMBO, but they all should be the same subset under the current single graph setting; meanwhile $(1,2,2), (1,1,2), (1,2,1), ...$ are valid subsets in COMBO, but they are invalid $k$-node combinations on a single graph. This limitation makes COMBO highly inefficient under this new problem setting, and therefore leads to inferior performance compared to our proposed method.

%% file: 13LargeGraph.tex
\section{Scalability on Large Graphs}\label{app large graph}
\begin{figure*}[ht]
    \begin{center}
        \centerline{\includegraphics[width=\textwidth]{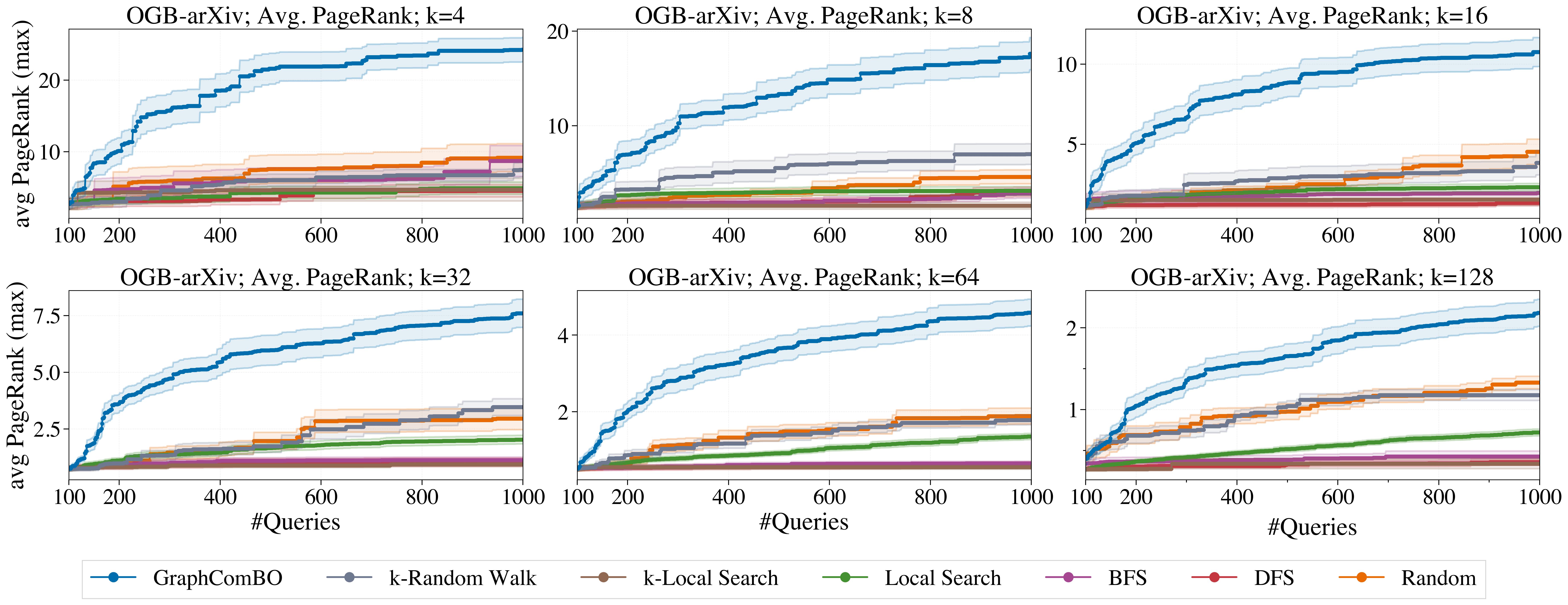}}
        \vspace{-0.2 cm}
        \caption{Maximizing avg. PageRank on the OGB-arXiv network ($|\mathcal{V}|=1.7\times10^5, \, |\mathcal{E}|=10^6$).} \label{fig ogb results}
    \end{center}
    \vspace{-0.2 cm}
\end{figure*}

\paragraph{Results on OGB-arXiv.} Since our framework assumes no prior knowledge of the full graph and takes a local modeling approach that gradually reveals the graph structure, it can scale to large underlying graphs with a reasonable choice of $k$. To better support this claim, we further test GraphComBO on a large social network OGB-arXiv ($|\mathcal{V}| = 1.7 \times 10^5$) from the open graph benchmark with $k$ up to 128, where the results in Figure~\ref{fig ogb results} show a clear advantage of our framework over the other baselines. Note that the local search methods underperform the random baseline under this setting, since exploration is relatively more important than exploitation.

\paragraph{Choice of $k$ in the experiments.} The subset size $k$ is set to $[2, 4, 8, 16, 32]$ across the experiments, which is a common paradigm in the literature of subset selection on graphs~\cite{kempe2003maximizing, leskovec2007cost, chen2016robust} with $k < 50$ (<1\% of the network), and the problem has been proven to be NP-hard in many problems due to the combinatorial explosion in search space $\binom{N}{k}$, e.g. $\binom{1000}{32} \approx 2.3 \times 10^{60}$. As such, the diminishing performance gain w.r.t. subset size $k$ poses a general challenge in the literature, and it becomes even more challenging in our setting, since the underlying function is fully black-boxed and we assume no prior information of the graph structure.

Nevertheless, the proposed method still generally outperforms the other baselines across all experiments, and in the least favorable case, it performs comparably to the local search, which is also a novel baseline introduced in our paper since it needs to operate on the proposed combo-graph.

%% file: 14NoisySetting.tex
\section{Settings under Noisy Observations}\label{app nosiy}
To show our framework’s capability of handling noise, we further conduct a noisy experiment at different noise levels on BA ($|\mathcal{V}| = 10k$) and WS ($|\mathcal{V}| = 1k$) networks with $k = 8$, where the goal is to maximize the average PageRank within a node subset, i.e., $f(\mathcal{S}) = \frac{1}{k} \sum_{i=1}^{k} PageRank(\mathcal{S}_i)$. with $\mathcal{S}$ being a subset of $k$ nodes $\{v_1, v_2, \ldots, v_k\}$ in the underlying graph $\mathcal{G}$.

\paragraph{A standardized underlying function with noise.} While it is difficult to show the noise level to the signal variance in real-world experiments because of the combinatorial space, we can construct a standardized signal under this synthetic setting with the following procedures. First, we standardized the PageRank scores over all nodes to mean=0 and std=1 in the original space (denoted as $PageRanks$). To standardize the underlying function in the combinatorial space, we multiply $\sqrt{k}$ to the average $PageRanks$ as the final underlying function $\tilde{f}(S)$, which is defined as follows:

\begin{equation}
    \tilde{f}(\mathcal{S}) = \sqrt{k} f_s(\mathcal{S}) = \frac{1}{\sqrt{k}} \sum_{i=1}^{k} PageRanks(S_i),
\end{equation}

where the expectation and variance of the transformed function $\tilde{f}(S)$ are:
\begin{align}
    \mathbb{E}[\tilde{f}(\mathcal{S})] & = 0 \\ 
    Var(\tilde{f}(\mathcal{S})) & = \frac{1}{k} Var\left(\sum_{i=1}^{k} PageRanks(\mathcal{S}_i)\right) \\ 
    & = \frac{1}{k} \times k \times Var(PageRanks(v)) \\ 
    & = 1
\end{align}

\begin{figure}[t]
    \begin{center}
        \centerline{\includegraphics[width=\textwidth]{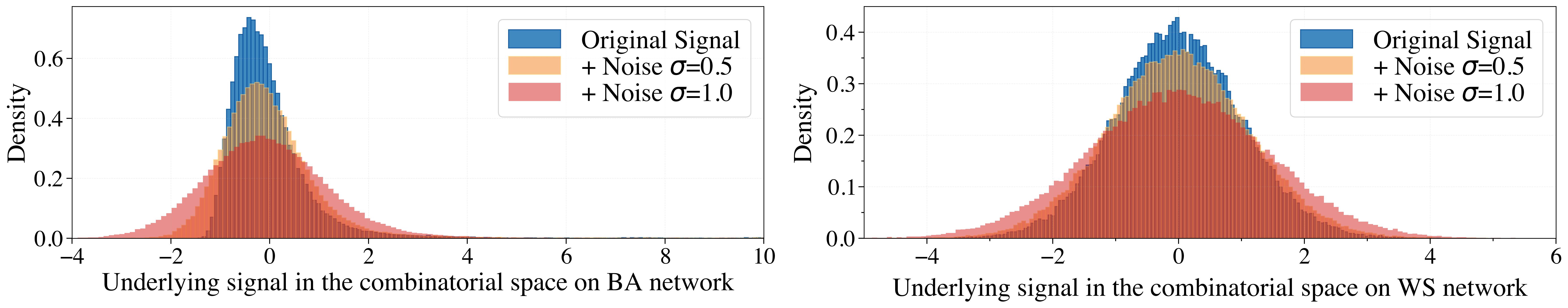}}
        \vspace{-0.2 cm}
        \caption{Density of underlying signals in the combinatorial space at different noise levels.} \label{fig noise hist}
    \end{center}
    \vspace{-0.2 cm}
\end{figure}

Now, we can simply add random Gaussian noise to $\tilde{f}(\mathcal{S})$. Specifically, we consider $\epsilon \sim N(0, \sigma_{\epsilon}^2)$ with $\sigma_{\epsilon}$ at $[0.1, 0.25, 0.5, 1]$, where the level of noise can be directly estimated since both the underlying function and noise now have a mean of $0$ and a standard deviation of $1$. In addition, we further plot the estimated density of the original and noisy signals in Figure~\ref{fig noise hist} to intuitively visualize the difference, which is done by randomly sampling $10^5$ observed values in the combinatorial space $\binom{N}{k}$.

\begin{figure}[t]
    \begin{center}
        \centerline{\includegraphics[width=\textwidth]{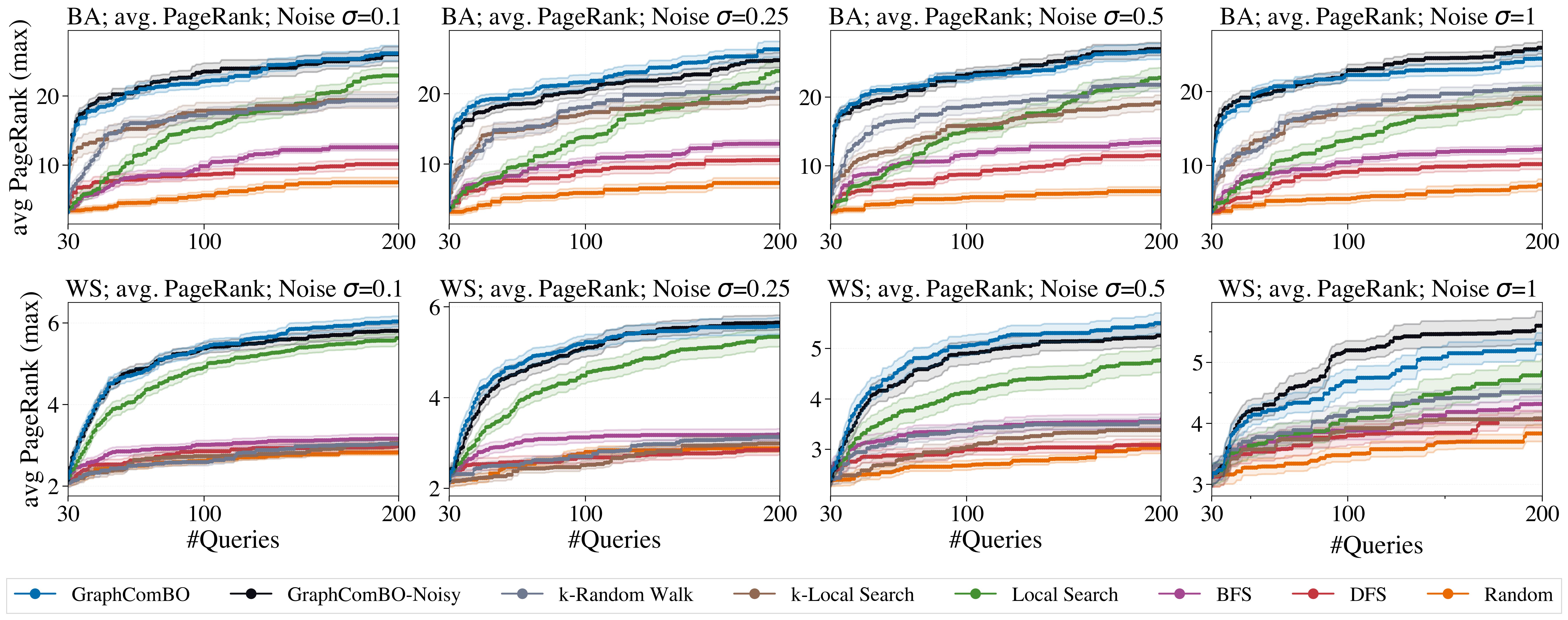}}
        \vspace{-0.2 cm}
        \caption{Maximizing avg. PageRank ($k=8$) on BA and WS at different noise levels.} \label{fig noisy results}
    \end{center}
    \vspace{-0.2 cm}
\end{figure}

\paragraph{GraphComBO-Noisy.} To better tackle the noisy observations, we implement \textit{\textbf{GraphComBO-Noisy}}, which uses the best posterior mean across both visited and non-visited combo-nodes within the combo-subgraph as the new center, and then compare its performance to the original method which is guided by the observation. The results in Figure~\ref{fig noisy results} show that the original method GraphComBO is robust to the noisy observations on both networks at different noise levels from $\sigma = 0.1$ to $\sigma = 1$. Compared with GraphComBO-Noisy, the observation-guided method performs comparably in most cases, except for a very noisy setting when $\sigma = 1$ on WS networks, where we can observe a clear advantage from the method guided by posterior mean, and it can be explained as follows.

Unlike classical discrete combinatorial functions of independent variables, the underlying functions in our problems are highly related to the graph structure. For example, BA networks are known for rich structural information due to the scale-free characteristics (i.e. node degree is exponentially distributed), which makes the distribution of the original signal heavily right-skewed with extreme values even after standardization (Figure~\ref{fig noise hist} Left). By contrast, the WS small-world network (randomly rewired from a ring) has more homogeneous node degrees, and thus the original signal will be more normally distributed after standardization (Figure~\ref{fig noise hist} Right). Therefore, the noise level (even at $\sigma = 1$) is less significant on BA networks when the algorithm finds the promising region, whereas on WS networks at $\sigma = 1$, just as the reviewer described, we can see the algorithm is ``misguided'' by the observations compared to the posterior mean when the signal is highly corrupted.

%% file: 15AblationStudies.tex
\section{Ablation Studies on Hyperparameters}\label{app ablation}
Lastly, we present the ablation study of BO's two hyperparameters: the combo-subgraph size $Q$ and the tolerance of continuous failures \texttt{failtol}, and analyze their influence on BO's performance when using the best query location as \texttt{restart\_method} over BA network and WS network. 

\begin{figure*}[t]
    \begin{center}
        \centerline{\includegraphics[width=\textwidth]{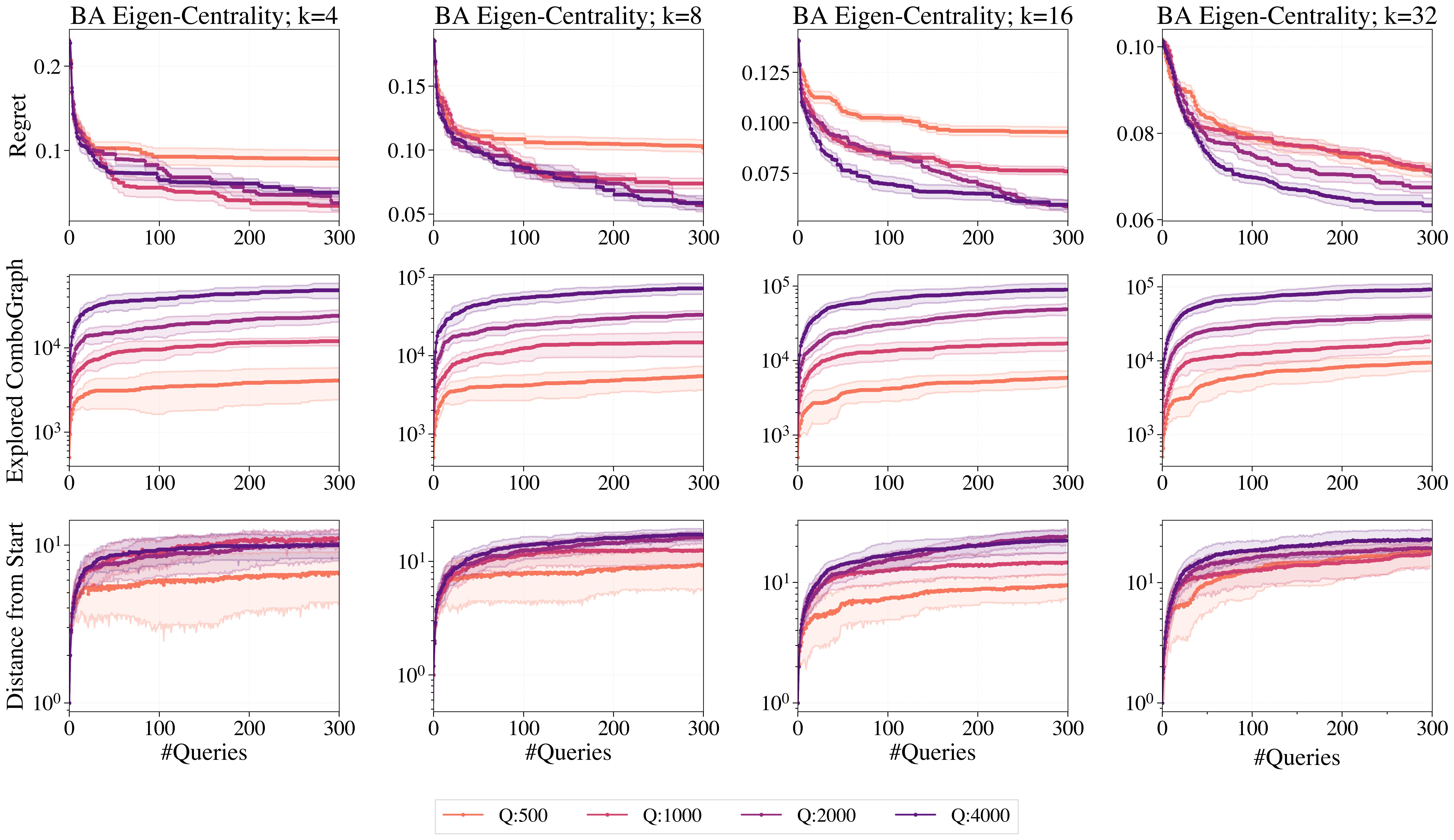}}  \vskip -0.2 cm
        \caption{Ablation study of $Q=[500,1k,2k,4k]$ with a fixed $\texttt{failtol}=30$ on BA network.} \label{fig ablation ba Q}
    \end{center}
    \vskip -0.5 cm
\end{figure*}

\begin{figure*}[t]
    \begin{center}
        \centerline{\includegraphics[width=\textwidth]{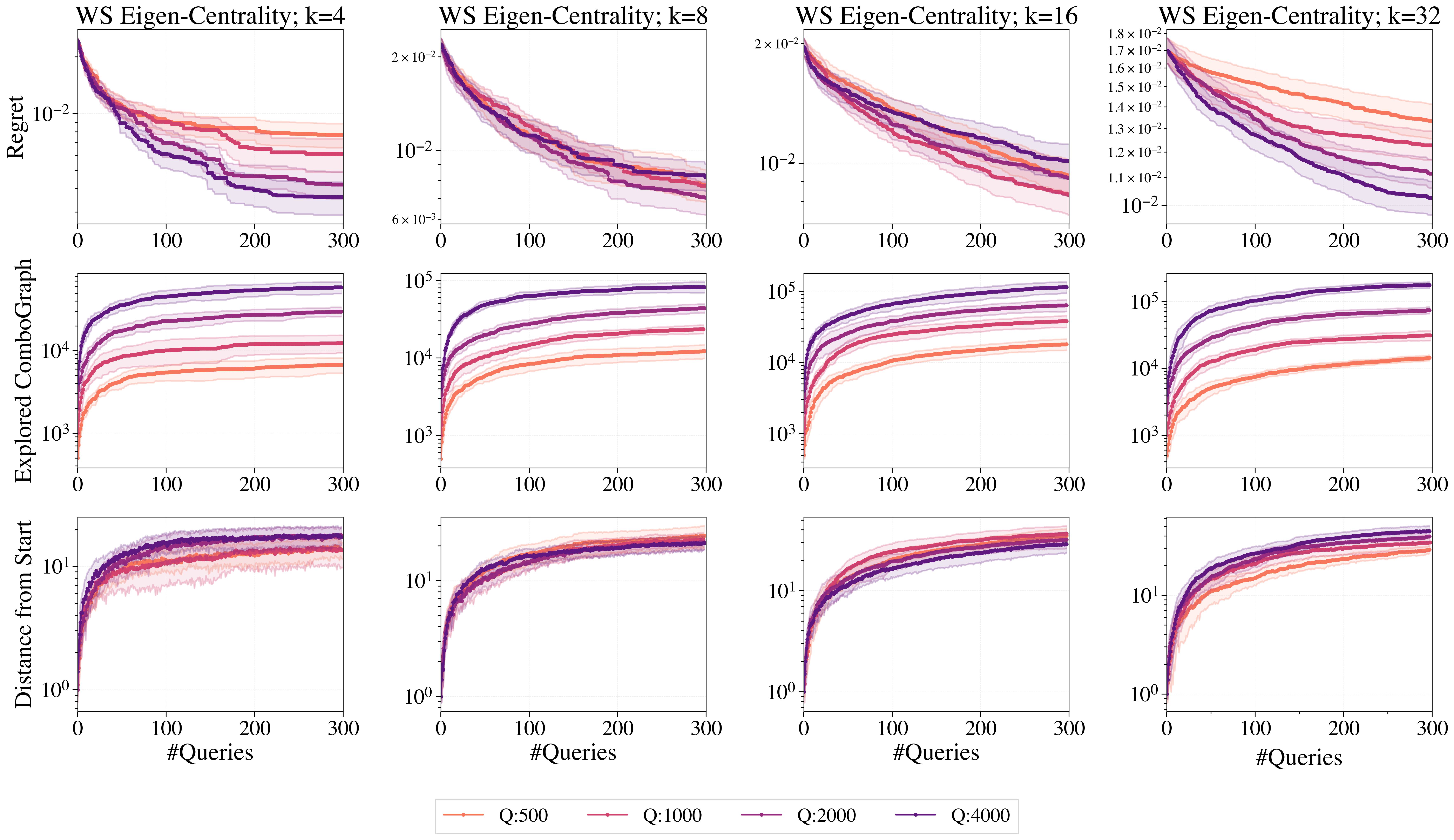}}  \vskip -0.2 cm
        \caption{Ablation study of $Q=[500,1k,2k,4k]$ with a fixed $\texttt{failtol}=30$ on WS network.} \label{fig ablation ws Q}
    \end{center}
    \vskip -0.5 cm
\end{figure*}

\paragraph{$\bf Q$}
To analyze the impact of $Q$ on BO's results, we fix $\texttt{failtol}=30$ and vary $Q$ at [500, 1000, 2000, 4000] and present the results in Figure \ref{fig ablation ba Q} for BA and Figure \ref{fig ablation ws Q} for WS. Overall, we can see that a larger $Q$ will lead to better performance in most situations, this is because we are starting with a random location in the combinatorial space and thus exploration is more important than exploitation. The explored combo-graph size and the distance of combo-subgraph center from the start also validate this interpretation, where a larger Q generally leads to a larger exploration region that contains more distant nodes of higher querying values, therefore leading to better search performance. In addition, we can also observe that as $k$ increases, the performance gain from larger $Q$ becomes more salient, which further corroborates the statements in Section \S\ref{sec experiment synthetic}.

\begin{figure*}[t]
    \begin{center}
        \centerline{\includegraphics[width=\textwidth]{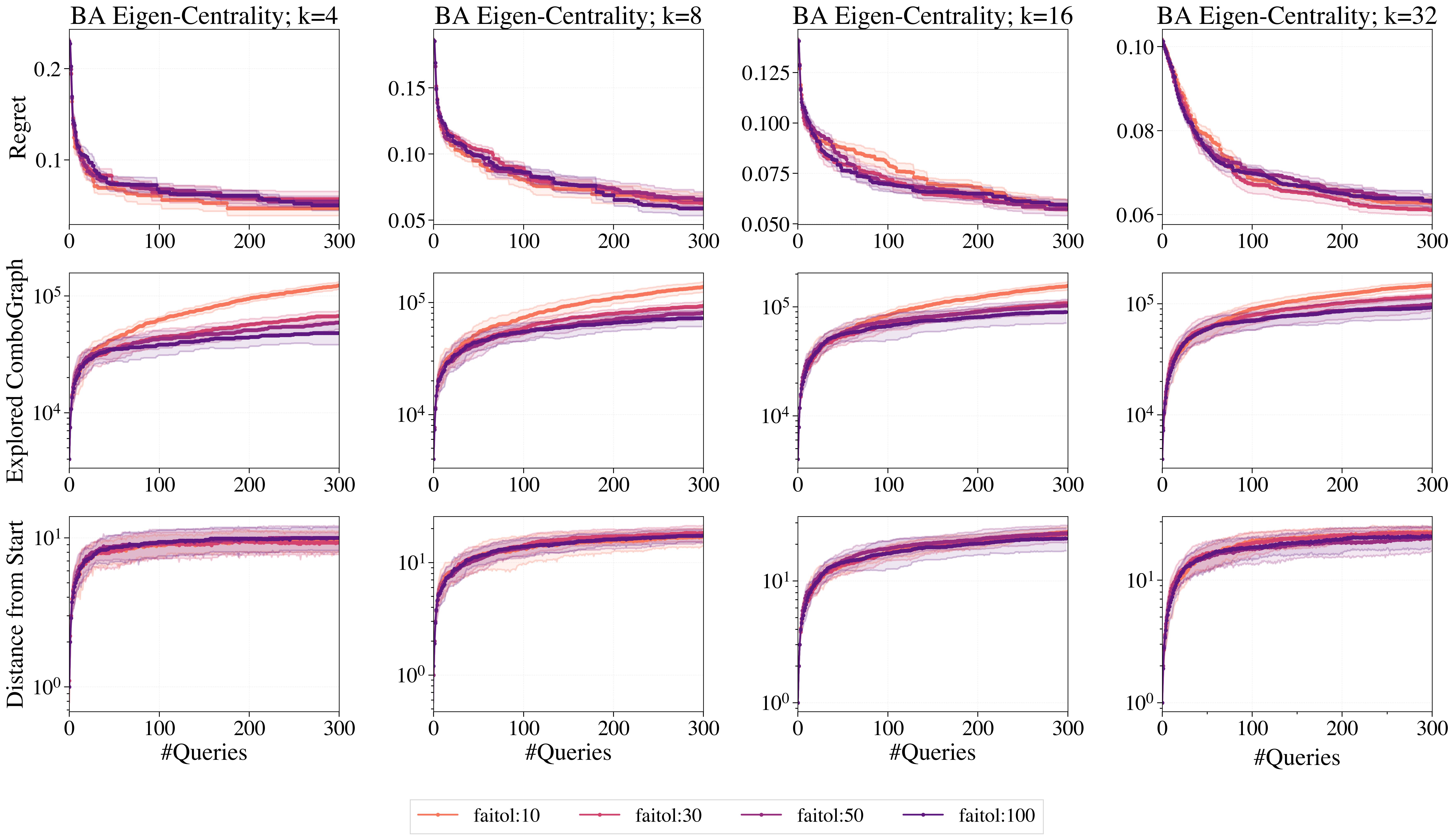}}
        \caption{Ablation study of $\texttt{failtol}=[10,30,50,100]$ with a fixed $Q=4000$ on BA network.} \label{fig ablation ba failtol}
    \end{center}
    \vskip -0.3 cm
\end{figure*}

\begin{figure*}[t]
    \begin{center}
        \centerline{\includegraphics[width=\textwidth]{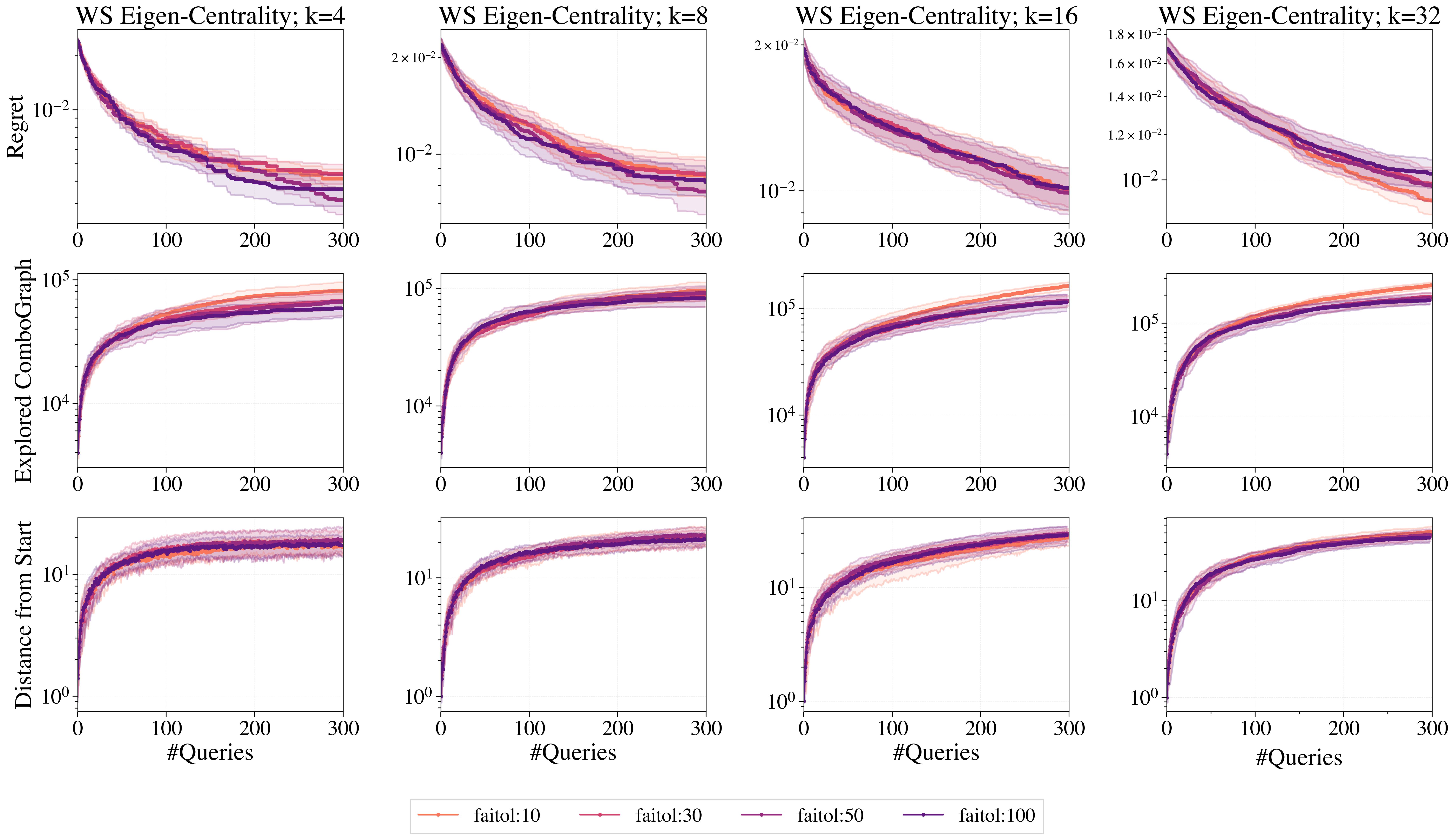}}
        \caption{Ablation study of $\texttt{failtol}=[10,30,50,100]$ with a fixed $Q=4000$ on WS network.} \label{fig ablation ws failtol}
    \end{center}
    \vskip -0.3 cm
\end{figure*}

\paragraph{\texttt{failtol}}
We analyze the influence of \texttt{failtol} on BO's performance by using a fixed $Q=4000$ and vary \texttt{failtol} at [10, 30, 50, 100], where the results are presented in Figure \ref{fig ablation ba failtol} for BA and Figure \ref{fig ablation ws failtol} for WS. We can observe that despite a small \texttt{failtol} is able to explore a larger region in the combinatorial space, the performance gain from this behavior is limited. We make the following explanations. Since the underlying functions used on both graphs (average eigenvector centralities) are rather smooth, restarting with a random location usually leads to worse overall search performance, and we adopt the best query location as our \texttt{restart\_method}. As a result, even if the algorithm restarts more frequently, it is still exploiting the regions around the same center. Nevertheless, we argue that with a different non-smooth underlying function, or when having a good initial location, a small \texttt{failtol} may have an advantage over the larger ones for its heavier exploitation behavior, and we leave this analysis to future work.


